\documentclass[letterpaper, 10 pt, conference]{class/ieeeconf}  % Comment this line out
\IEEEoverridecommandlockouts                              % This command is only
\usepackage{xcolor}
\usepackage{siunitx}  
\usepackage{graphicx}
 \graphicspath{./figures/robot_demonstration}
\usepackage{amsmath}
\usepackage{amssymb}
\usepackage{latexsym}
\usepackage{url}
\usepackage{cite}
\usepackage{relsize}
\usepackage{multirow}
\usepackage{afterpage}
\usepackage{ifthen}

\usepackage{graphicx}
\usepackage{algpseudocode,algorithm,algorithmicx}
\usepackage{flushend}
\usepackage{epstopdf}
\usepackage{stackengine}
\usepackage{textcomp} % new
\usepackage{makecell}
\usepackage{gensymb}
\usepackage{booktabs}
\usepackage{makecell}
\usepackage{hhline}
\usepackage{varwidth}
\usepackage{courier}
\usepackage{lipsum}
\usepackage{xspace}
\usepackage{booktabs}
\usepackage{diagbox}
\usepackage{pifont}
\usepackage{xcolor} % new
\newcommand{\xmark}{\ding{55}}

\makeatletter
\let\NAT@parse\undefined
\makeatother
\usepackage[bookmarks=false, linkcolor=blue, urlcolor=blue, citecolor=blue]{hyperref} 
\hypersetup{
    colorlinks=true,
    linkcolor=red,
    filecolor=magenta,      
    urlcolor=blue,
    pdfstartview={FitH},
    citecolor =blue
    }

\usepackage{xspace}

\title{\LARGE \bf HabiCrowd: A High Performance Simulator for Crowd-Aware \\Visual Navigation}

\author{An Vuong$^{1}$, Toan Nguyen$^{1}$, Minh Nhat Vu$^{2,3,*}$, Baoru Huang$^4$, H.T.T Binh$^5$, Thieu Vo$^6$, Anh Nguyen$^7$
\thanks{$^1$ FPT Software AI Center, Vietnam {\tt anvd2@fpt.com}}
\thanks{$^2$ Automation \& Control Institute, TU Wien, Austria %{\tt vu@acin.tuwien.ac.at}
}
\thanks{$^3$ Austrian Institute of Technology (AIT), Austria}
\thanks{$^4$ Imperial College London, UK
}
\thanks{$^5$ Hanoi University of Science and Technology, Vietnam 
}
\thanks{$^6$ Faculty of Mathematics and Statistics, TDTU, Vietnam} %{\tt vongocthieu@tdtu.edu.vn}}
\thanks{$^7$ Department of Computer Science, University of Liverpool, UK %{\tt anh.nguyen@liverpool.ac.uk}
}
\thanks{$^*$ Corresponding author {\tt minh.vu@ait.ac.at}
}
}

\begin{document}
% Macros

\newtheorem{problem}{Problem}
\newtheorem{lemma}{Lemma}
\newtheorem{theorem}[lemma]{Theorem}
\newtheorem{claim}{Claim}
\newtheorem{corollary}[lemma]{Corollary}
\newtheorem{definition}[lemma]{Definition}
\newtheorem{proposition}[lemma]{Proposition}
\newtheorem{remark}[lemma]{Remark}
\newenvironment{LabeledProof}[1]{\noindent{\it Proof of #1: }}{\qed}

\def\beq#1\eeq{\begin{equation}#1\end{equation}}
\def\bea#1\eea{\begin{align}#1\end{align}}
\def\beg#1\eeg{\begin{gather}#1\end{gather}}
\def\beqs#1\eeqs{\begin{equation*}#1\end{equation*}}
\def\beas#1\eeas{\begin{align*}#1\end{align*}}
\def\begs#1\eegs{\begin{gather*}#1\end{gather*}}

\newcommand{\poly}{\mathrm{poly}}
\newcommand{\eps}{\epsilon}
\newcommand{\e}{\epsilon}
\newcommand{\polylog}{\mathrm{polylog}}
\newcommand{\rob}[1]{\left( #1 \right)} %Round Brackets
\newcommand{\sqb}[1]{\left[ #1 \right]} %square Brackets
\newcommand{\cub}[1]{\left\{ #1 \right\} } %curly brackets
\newcommand{\rb}[1]{\left( #1 \right)} %Round
\newcommand{\abs}[1]{\left| #1 \right|} %| |
\newcommand{\zo}{\{0, 1\}}
\newcommand{\zonzo}{\zo^n \to \zo}
\newcommand{\zokzo}{\zo^k \to \zo}
\newcommand{\zot}{\{0,1,2\}}
\newcommand{\en}[1]{\marginpar{\textbf{#1}}}
\newcommand{\efn}[1]{\footnote{\textbf{#1}}}
\newcommand{\vecbm}[1]{\boldmath{#1}} %more general (handles greek letters)
\newcommand{\uvec}[1]{\hat{\vec{#1}}}
\newcommand{\thv}{\vecbm{\theta}}
\newcommand{\junk}[1]{}
\newcommand{\var}{\mathop{\mathrm{var}}}
\newcommand{\rank}{\mathop{\mathrm{rank}}}
\newcommand{\diag}{\mathop{\mathrm{diag}}}
\newcommand{\tr}{\mathop{\mathrm{tr}}}
\newcommand{\acos}{\mathop{\mathrm{acos}}}
\newcommand{\atantwo}{\mathop{\mathrm{atan2}}}
\newcommand{\SVD}{\mathop{\mathrm{SVD}}}
\newcommand{\quadf}{\mathop{\mathrm{q}}}
\newcommand{\linterp}{\mathop{\mathrm{l}}}
\newcommand{\sgn}{\mathop{\mathrm{sign}}}
\newcommand{\sym}{\mathop{\mathrm{sym}}}
\newcommand{\avg}{\mathop{\mathrm{avg}}}
\newcommand{\mean}{\mathop{\mathrm{mean}}}
\newcommand{\erf}{\mathop{\mathrm{erf}}}
\newcommand{\grad}{\nabla}
\newcommand{\R}{\mathbb{R}}
\newcommand{\defeq}{\triangleq}
\newcommand{\dims}[2]{[#1\!\times\!#2]}
\newcommand{\sdims}[2]{\mathsmaller{#1\!\times\!#2}}
\newcommand{\udims}[3]{#1}
\newcommand{\udimst}[4]{#1}
\newcommand{\com}[1]{\rhd\text{\emph{#1}}}
\newcommand{\ind}{\hspace{1em}}
\newcommand{\argmin}[1]{\underset{#1}{\operatorname{argmin}}}
\newcommand{\floor}[1]{\left\lfloor{#1}\right\rfloor}
\newcommand{\step}[1]{\vspace{0.5em}\noindent{#1}}
\newcommand{\quat}[1]{\ensuremath{\mathring{\mathbf{#1}}}}
\newcommand{\norm}[1]{\left\lVert#1\right\rVert}
\newcommand{\ignore}[1]{}
\newcommand{\specialcell}[2][c]{\begin{tabular}[#1]{@{}c@{}}#2\end{tabular}}
\newcommand*\Let[2]{\State #1 $\gets$ #2}
\newcommand{\algorithmicbreak}{\textbf{break}}
\newcommand{\Break}{\State \algorithmicbreak}
\newcommand{\ra}[1]{\renewcommand{\arraystretch}{#1}}

\renewcommand{\vec}[1]{\mathbf{#1}} %looks better

\algdef{S}[FOR]{ForEach}[1]{\algorithmicforeach\ #1\ \algorithmicdo}
\algnewcommand\algorithmicforeach{\textbf{for each}}
\algrenewcommand\algorithmicrequire{\textbf{Require:}}
\algrenewcommand\algorithmicensure{\textbf{Ensure:}}
\algnewcommand\algorithmicinput{\textbf{Input:}}
\algnewcommand\INPUT{\item[\algorithmicinput]}
\algnewcommand\algorithmicoutput{\textbf{Output:}}
\algnewcommand\OUTPUT{\item[\algorithmicoutput]}

\maketitle
\thispagestyle{empty}
\pagestyle{empty}

%%%%%%%%%%%%%%%%%%%%%%%%%%%%%%%%%%%%%%%%%%%%%%%%%%%%%%%%%%%%%%%%%%%%%%%%%%%%%%%%
\begin{abstract}
Visual navigation, a foundational aspect of Embodied AI (E-AI) and robotics has been extensively studied in the past few years. While many 3D simulators have been introduced for the visual navigation tasks, scarcely works have combined human dynamics, creating the gap between simulation and real-world applications. Furthermore, current 3D simulators incorporating human dynamics have several limitations, particularly in terms of computational efficiency, which is a promise of modern simulators. To overcome these issues, we introduce HabiCrowd, the new standard benchmark for crowd-aware visual navigation that includes a crowd dynamics model with diverse human settings into photorealistic environments. Empirical evaluations demonstrate that our proposed human dynamics model achieves state-of-the-art performance in collision avoidance while exhibiting superior computational efficiency compared to its counterparts. We leverage HabiCrowd to conduct several comprehensive studies on crowd-aware visual navigation tasks and human-robot interactions. The source code and data can be found at \href{https://habicrowd.github.io/}{https://habicrowd.github.io/}.
\end{abstract}

%%%%%%%%%%%%%%%%%%%%%%%%%%%%%%%%%%%%%%%%%%%%%%%%%%%%%%%%%%%%%%%%%%%%%%%%%%%%%%%%

\section{Introduction}
Embodied AI (E-AI), the intersection between computer vision, machine learning, and robotics, has gained interest amongst scientists in the past few years~\cite{yadav2023ovrl}. Training E-AI agents that can perceive, reason, and
interact with the environment offers significant potential for sim2real applications~\cite{srivastava2022behavior}. The progress in E-AI research has been rapidly accelerated~\cite{yadav2023habitat} thanks to the developments of fast and high-fidelity 3D simulators~\cite{wijmans2019dd} that enable large-scale computational parallelization~\cite{chaplot2020object}. Among all E-AI studies, one of the most fundamental problems is visual navigation~\cite{gupta2017cognitive}, which involves training an agent to navigate to a given goal using perception from the sensory inputs~\cite{ramakrishnan2021hm3d}. While many benchmarks have been proposed to address the visual navigation tasks~\cite{chaplot2020object, bansal2020combining}, most of them assume that navigating environments are static and do not consider human dynamics. Conversely, agents typically interact within dynamic environments, particularly those featuring humans whose positions constantly change~\cite{liu2021decentralized}. If robots are to efficiently assist humans in routine duties, they must possess the capability to navigate effectively in \textit{crowded and dynamic environments}~\cite{monaci2022dipcan}. Therefore, human dynamics is an essential factor that needs to be thoroughly examined when developing E-AI benchmarks~\cite{bonin2008visual}. This paper examines visual navigation in the presence of human dynamics to bridge the gap between simulation and real-life scenarios.

\begin{figure*}[ht!]
  \centering
  \setlength{\tabcolsep}{2pt} % Adjust the space between images
  \begin{tabular}{ccccc}
    \shortstack{\includegraphics[width=0.19\linewidth]{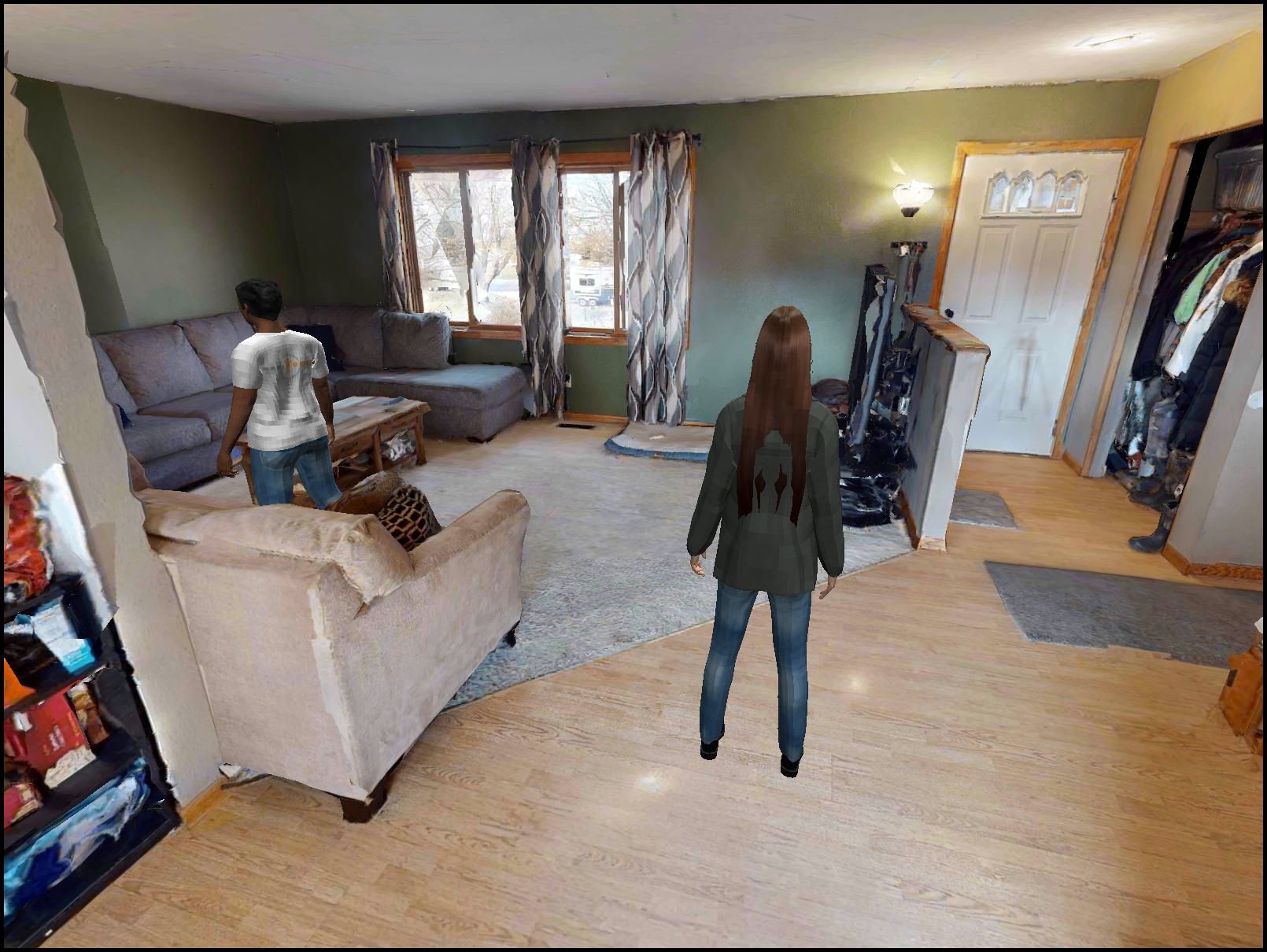}} &
    \shortstack{\includegraphics[width=0.19\linewidth]{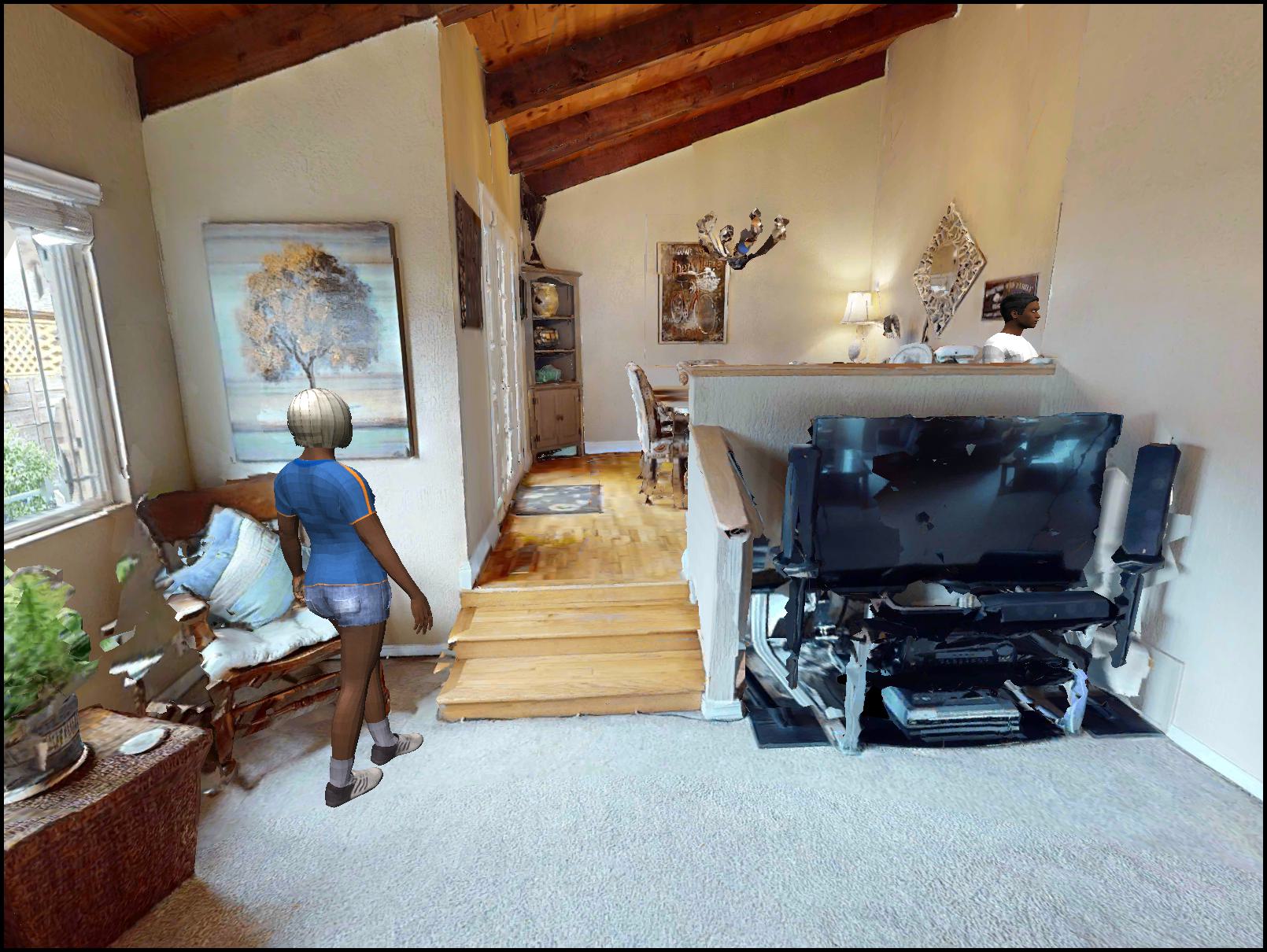}} &
    \shortstack{\includegraphics[width=0.19\linewidth]{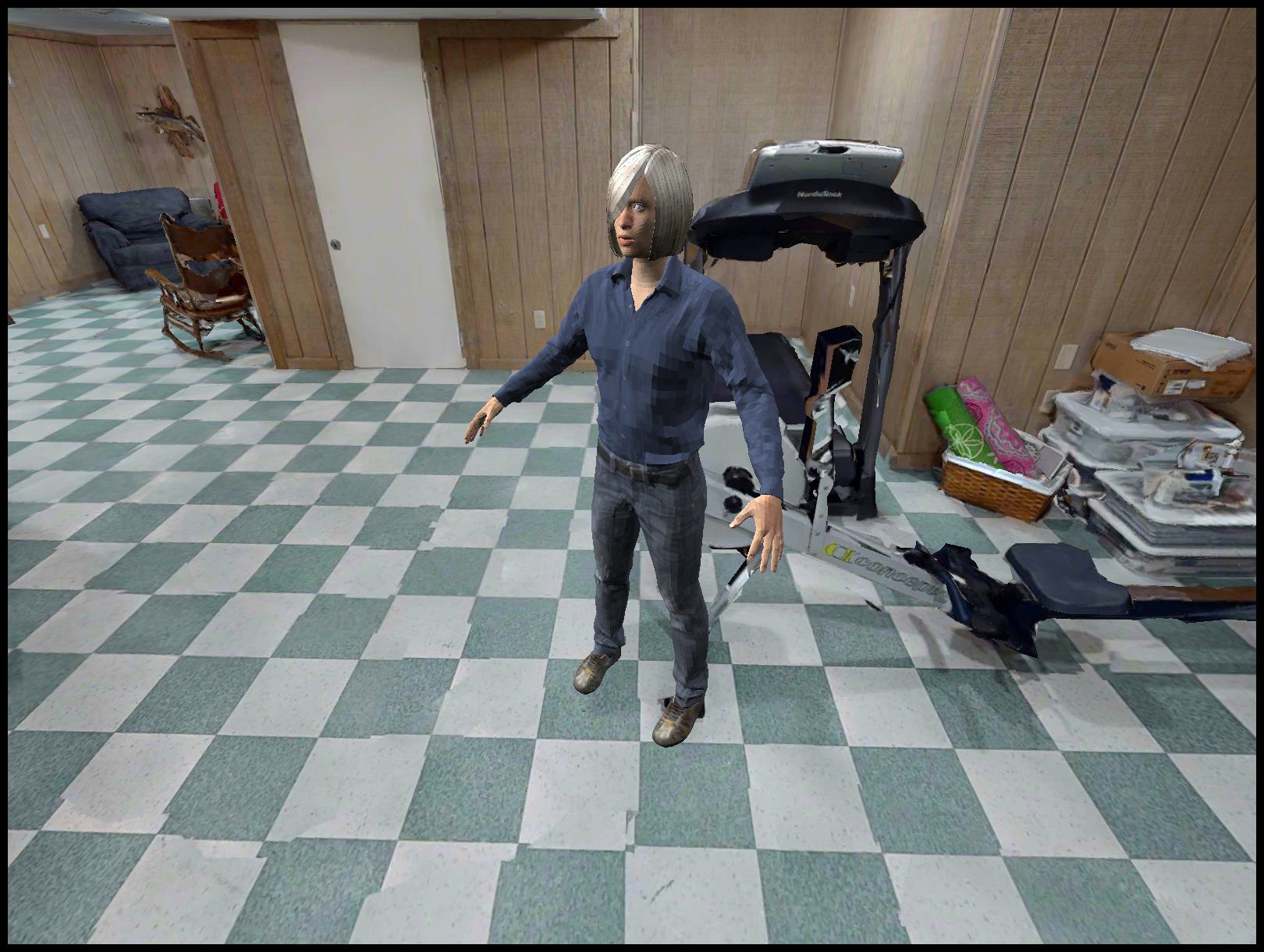}} &
    \shortstack{\includegraphics[width=0.19\linewidth]{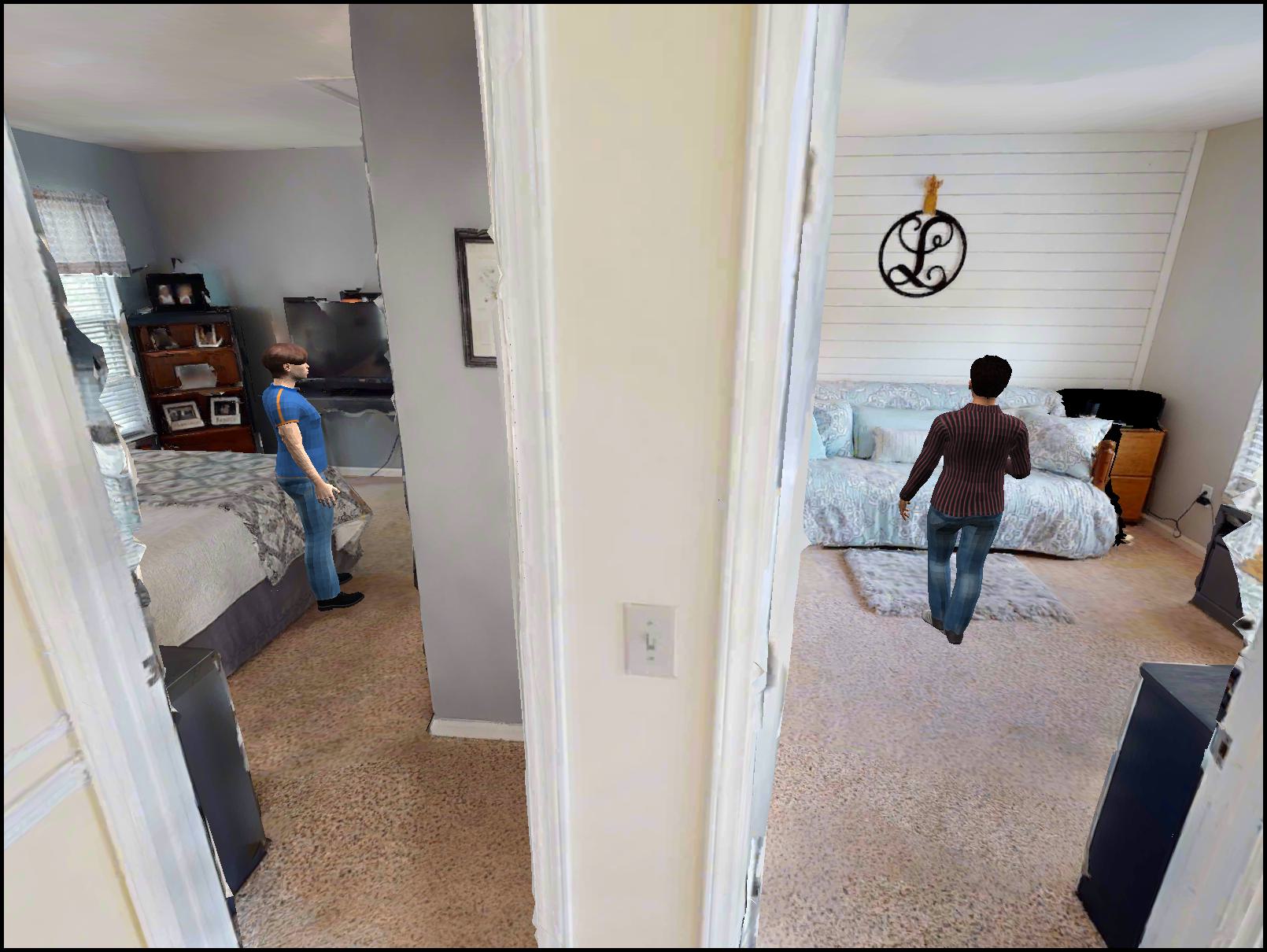}} &
    \shortstack{\includegraphics[width=0.19\linewidth]{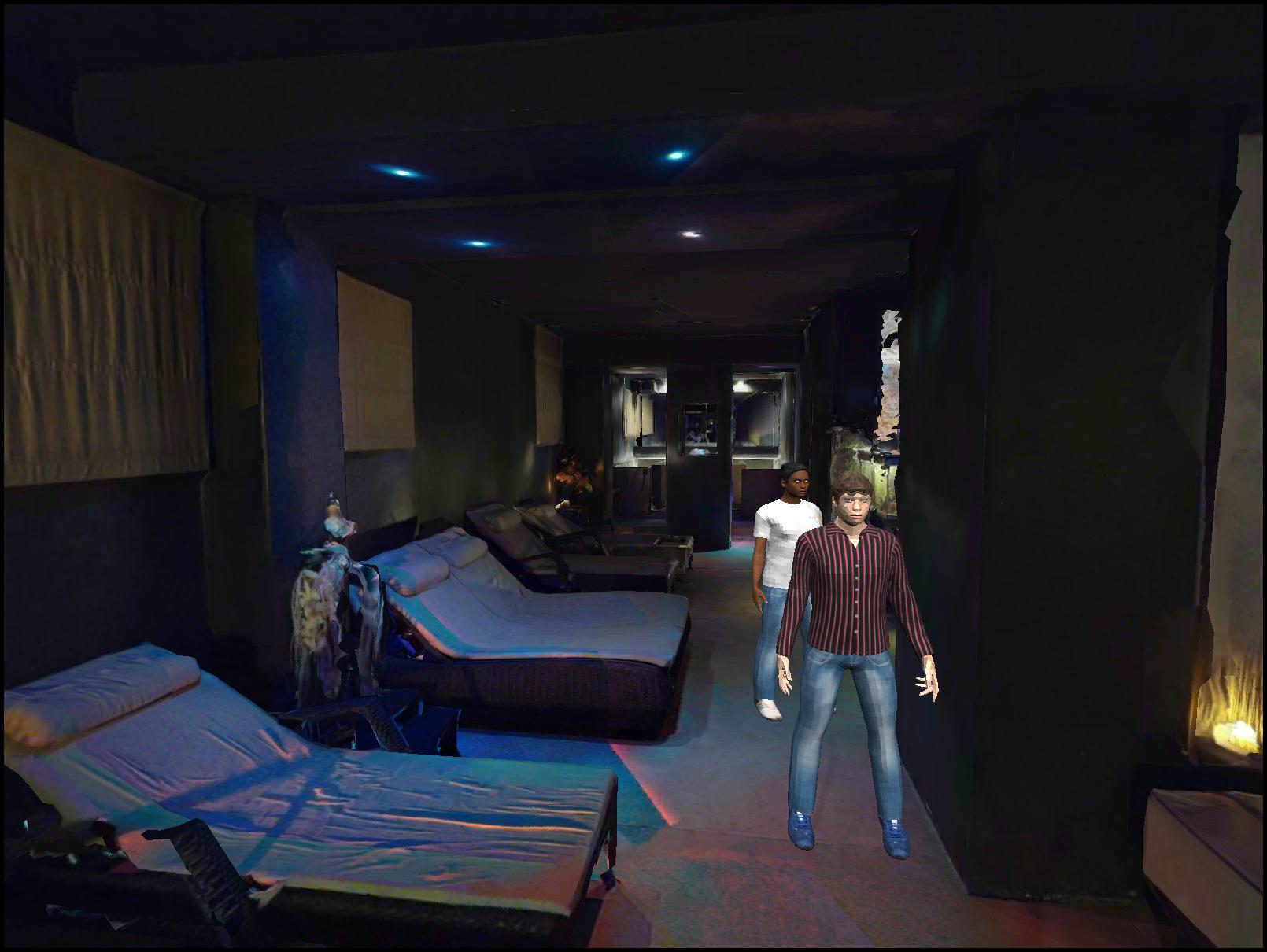}} \\
    % Adjusting the gap between the rows of images if necessary
    \shortstack{\includegraphics[width=0.19\linewidth]{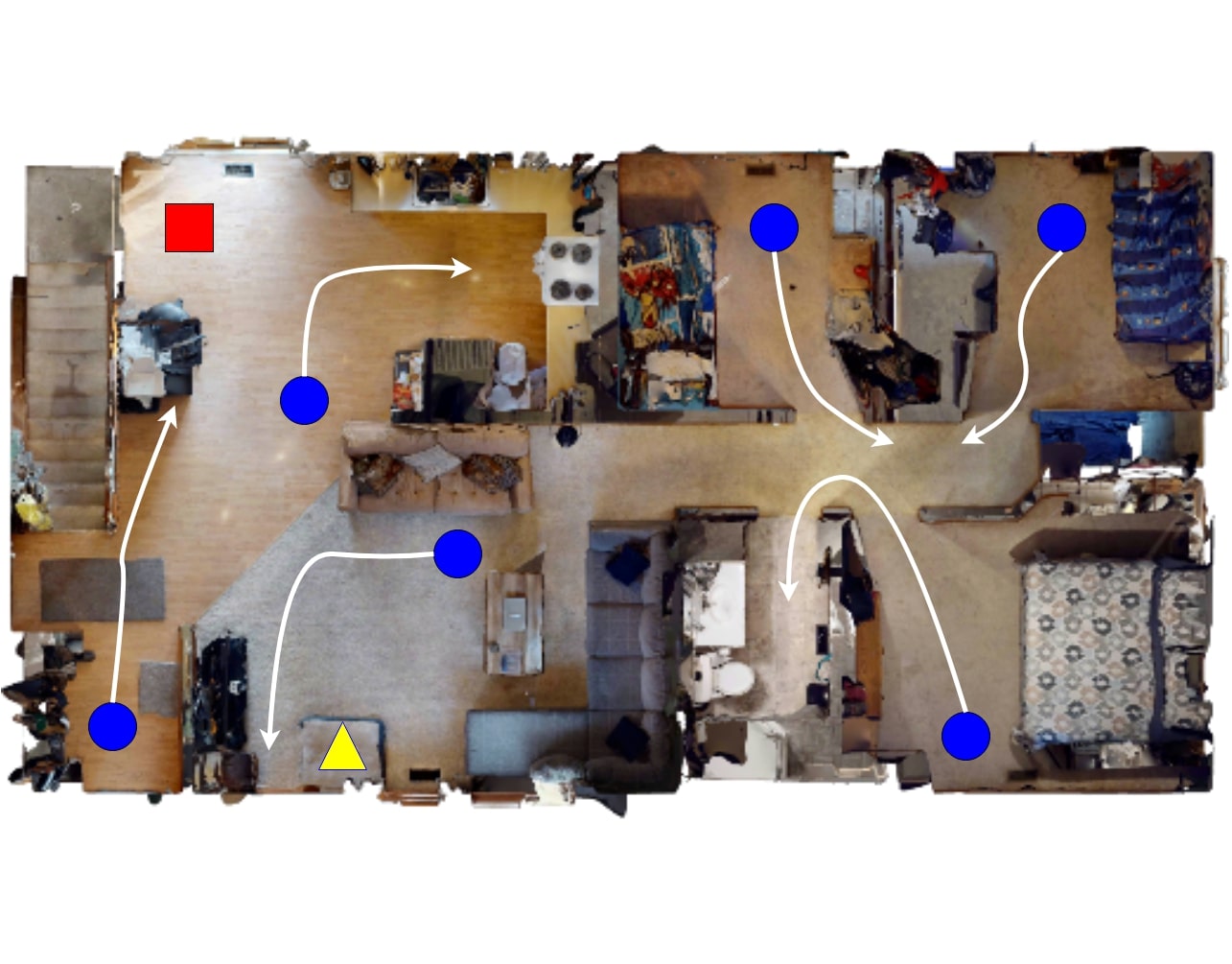}} &
    \shortstack{\includegraphics[width=0.19\linewidth]{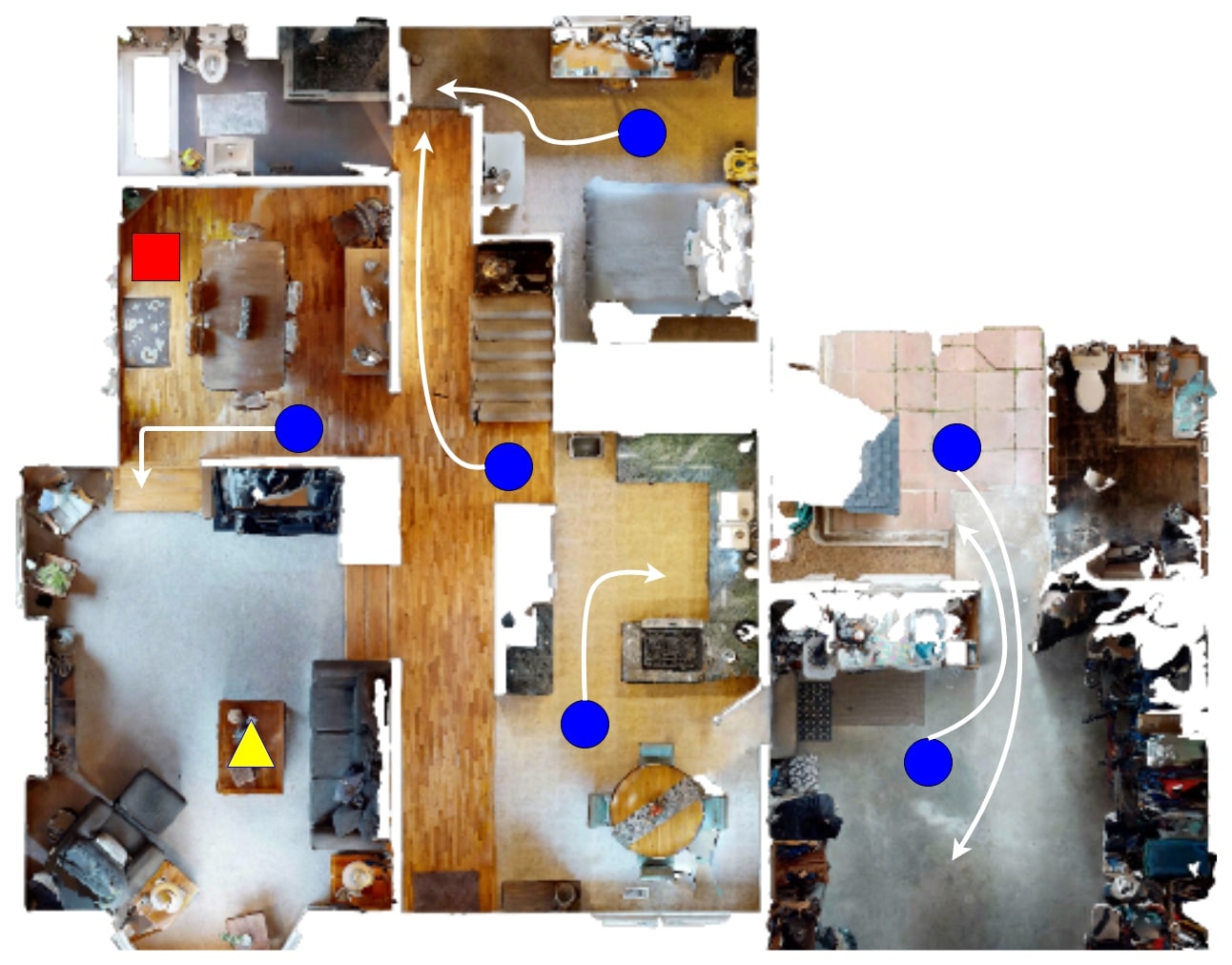}} &
    \shortstack{\includegraphics[width=0.19\linewidth]{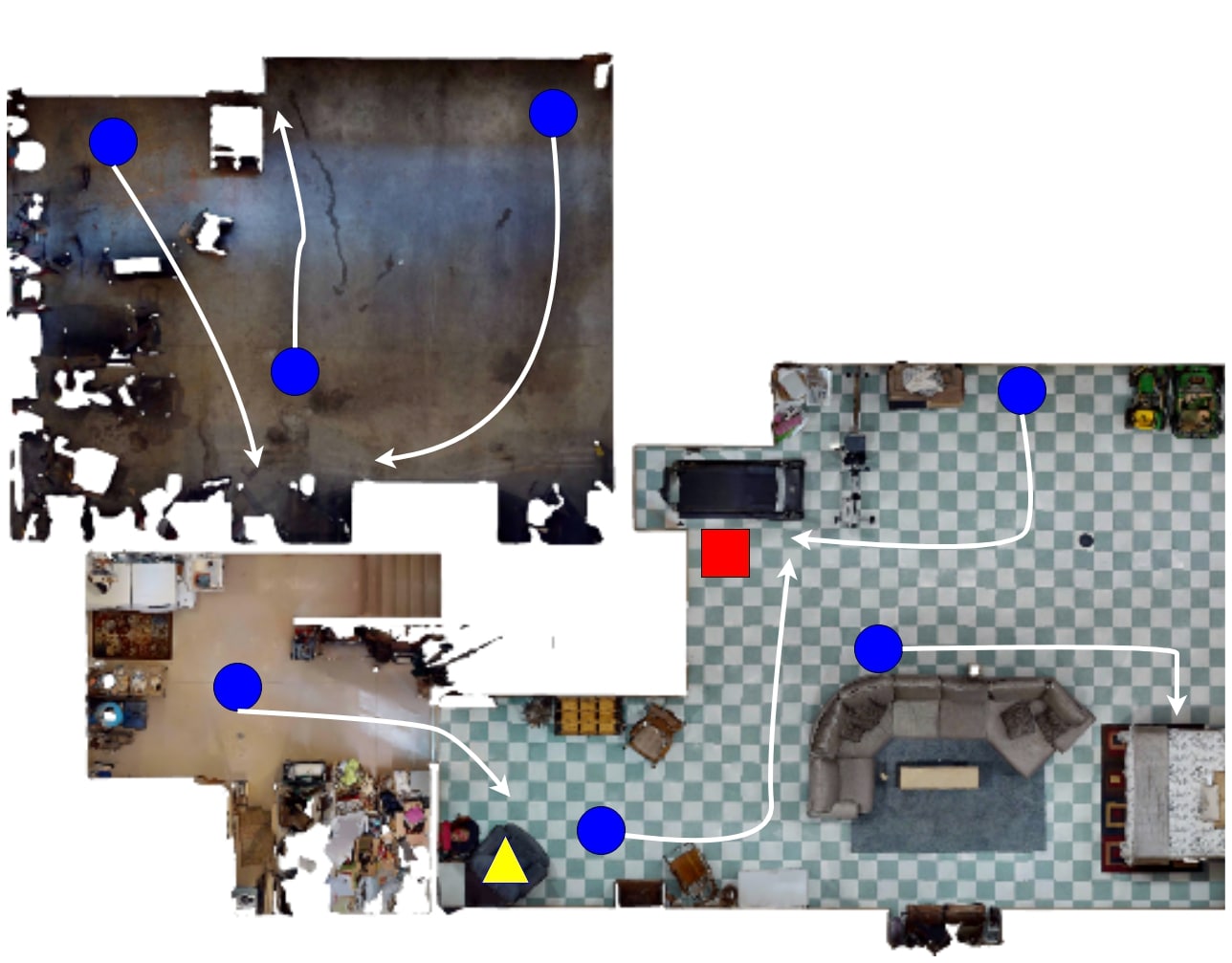}} &
    \shortstack{\includegraphics[width=0.19\linewidth]{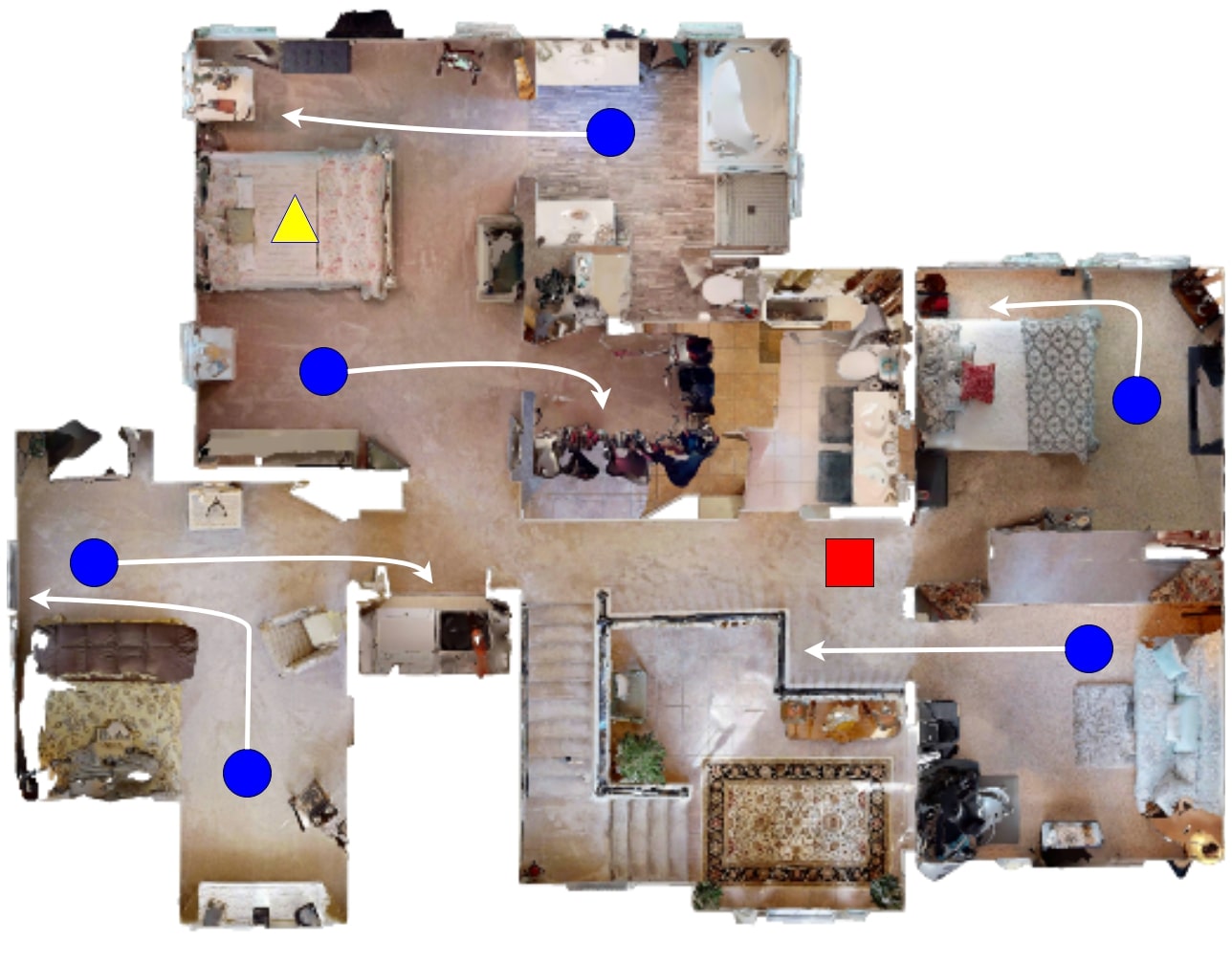}} &
    \shortstack{\includegraphics[width=0.19\linewidth]{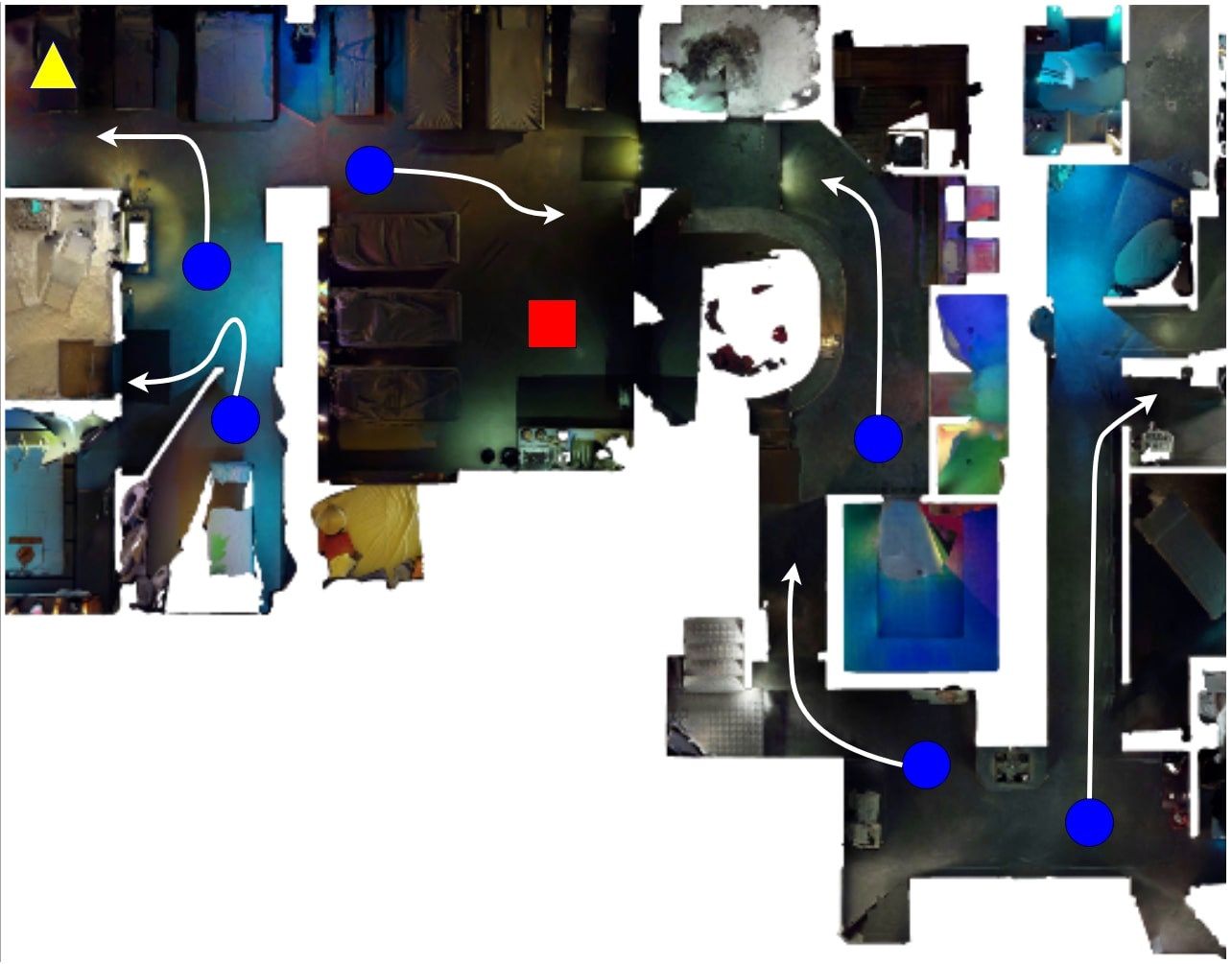}} \\
  \end{tabular}
  \vspace{1ex}
  \caption{We present HabiCrowd, a new benchmark for crowd-aware visual navigation. The top row shows some example scenes, and the bottom row shows the floor plan with human dynamics.}
  \label{fig:intro}
\end{figure*}

The visual navigation task with the inclusion of human dynamics is widely referred to as the crowd-aware visual navigation task in the robotics community~\cite{chen2019crowd}. This task is challenging because the agent needs to learn the patterns of human actions~\cite{kretzschmar2016socially}, which are complex and unstructured~\cite{dugas2022navdreams}. Furthermore, it is infeasible to predict the motion dynamics in crowded habitats as the motion of each person is typically affected by their neighbours~\cite{alahi2016social}.  Most prior works train agents using 2D simulators~\cite{chen2017socially,liu2021decentralized}, which is impractical since robots perceive human activities via visual sensors rather than mathematical social force models~\cite{dugas2022navdreams}.

Recent attempts to add dynamic humans to 3D photorealistic simulators~\cite{2022igibsonchallenge, makoviychuk2021isaac} have shown limitations in human dynamics models and world scale. For instance, the limited number of humans employed in~\cite{2022igibsonchallenge}  can hinder generalization in human recognition~\cite{alahi2016social}. Additionally, one of the most critical drawbacks of recent 3D simulators integrating human dynamics lies in their computational efficiency~\cite{yokoyama2022benchmarking}, even though E-AI simulators strive to facilitate rapid rendering and parallelization~\cite{wijmans2019dd}. In general, the lack of 3D simulators with well-designed human dynamics reduces the applicability of developed methods, highlighting the need for a standardized 3D benchmark that integrates human behaviors.

Motivated by these shortcomings,  we introduce \textbf{HabiCrowd}, a new benchmark for the crowd-aware visual navigation task. HabiCrowd includes comprehensive social force-based human dynamics with diverse density settings operating across numerous high-fidelity scenes from the Habitat-Matterport 3D dataset (HM3D)~\cite{ramakrishnan2021hm3d}. To assess the performance of HabiCrowd, several experiments are conducted, including benchmarking the human dynamics model, rendering speed, and memory utilization. Experimental results reveal that HabiCrowd achieves collision-free while running significantly faster than the state-of-the-art human dynamics model. Additionally, our HabiCrowd also exhibits a remarkable advantage in rendering speed and memory utilization compared to other related benchmarks. Upon HabiCrowd, we propose a new metric for crowd-aware navigation and benchmark two visual navigation tasks. The outcomes highlight the importance of human dynamics in navigating procedures. We also demonstrate that our new simulator enables various studies regarding human density and human-robot interactions. Our contributions are summarized as follows:
%We further develop a strong baseline called DaViT (\textbf{D}ouble \textbf{A}ttention \textbf{Vi}sion \textbf{T}ransformer) that utilizes the self-attention mechanism to exploit the graphical features from the observations to effectively avoid collisions with humans.

\begin{itemize}
\item We present HabiCrowd, a new dataset and benchmark for crowd-aware visual navigation. Experiments show that our simulator excels in terms of performance, human diversity and computational efficiency.

\item We comprehensively benchmark navigation tasks using HabiCrowd to showcase that our simulator offers detailed analyses, such as studies on human density and human-robot avoidance in crowd-aware environments.
\end{itemize}
 \section{Related Works}
%In~\cite{}, the authors trained MPC with an end-to-end deep neural network based on raw images as the input. Hirose~\textit{et al.}~\cite{} addressed the problem of finding a navigation path while avoiding collisions with undiscovered objects by using a deep MPC model.
%For instance, Zhu~\textit{et al.}~\cite{zhu2017target} proposed an approach based on the actor-critic method to give a robust ability to generalize in different environments while the authors in~\cite{kahn2018self} combined model-free and model-based methods to obtain a computation graph for optimal navigation.
\textbf{Visual Navigation.}
Traditionally, roboticists have addressed visual navigation tasks by utilizing control methods such as Model Predictive Control (MPC)~\cite{hirose2019deep}. Several following works have been proposed to improve control-based methods~\cite{li2015vision}. The idea of control methods has been conserved by utilizing the advantage of deep learning~\cite{hirose2019deep}. Nevertheless, control-based methods demand a dynamics model and are often saturated when the number of ambiguities and the prediction horizon grows~\cite{lucia2020stability}. Recently, achievements in reinforcement learning (RL)~\cite{mnih2015human} have brought about a significant shift in visual navigation~\cite{yang2018visual}. Over the years, deep RL approaches have emerged as a widely adopted solution for addressing visual navigation tasks~\cite{ yadav2023ovrl, khandelwal2022simple}, and many RL-based methods have achieved state-of-the-art results in visual navigation tasks such as point-goal navigation~\cite{wijmans2019dd}, object-goal navigation~\cite{ramrakhya2023pirlnav}, image-goal navigation~\cite{yadav2023ovrl}. %Fascinating by these achievements, we benchmark state-of-the-art RL baselines upon our proposed simulator.

\textbf{Crowd-aware Navigation.}
Navigation in crowded circumstances has significantly been examined by the robotics community~\cite{chen2019crowd}. Earlier methods in crowd-aware navigation are primarily based on predicting the trajectories of human entities~\cite{ferrer2013robot}. As the number of humans grows, the environment becomes exponentially denser; consequently, these trajectory-based approaches often suffer from the well-known freezing robot problem~\cite{trautman2010unfreezing}. Learning-based methods have been introduced to overcome the freezing robot issue by automatically determining optimal waypoints~\cite{liu2021decentralized}. Following this idea, RL frameworks are widely adopted to perceive human behaviors and interactions in a latent mechanism~\cite{chen2019crowd,trautman2020real}. However, RL methods based on 2D simulators often suppose that the dynamics of all human entities are explicit and well-determined while the real-life interactions and human dynamics are much more arbitrary and complex~\cite{trautman2020real}. To address this limitation, we study crowd-aware navigation in 3D settings, wherein the agent can only perceive the environment through egocentric inputs instead of 2D floor plans with known human dynamics.

 %; however, most works only employ 2D simulators or small-scale experiment setups~\cite{kruse2013human}.%Nonetheless, since the computation demand is astronomical, M%can be separated into two stages: predicting the trajectories of human entities, and planning a human avoidance route for robots~\cite{borenstein1989real, bennewitz2005learning, ferrer2013robot}. 

\textbf{3D Simulators for Visual Navigation.}
There have been a large number of simulators for the tasks of visual navigation~\cite{duan2022survey}. Gazebo, introduced in~\cite{koenig2004design}, is one of the first steps towards virtualizing the real world. Followed by Gazebo, several simulators have been established to study daily-conventional behaviors of robots, such as UnrealCV~\cite{qiu2017unrealcv}, AI2-THOR~\cite{kolve2017ai2}, Gibson~\cite{xia2018gibson}, Habitat~\cite{szot2021habitat}. Although the mentioned simulators provide opportunities to study
interactions with humans, none of them have yet included human dynamics until the establishment of iGibson-Social Navigation (iGibson-SN)~\cite{2022igibsonchallenge}. More recently, Isaac Sim~\cite{makoviychuk2021isaac} is also established with the inclusion of virtual humans to the scenes. Nevertheless, the main drawbacks of both iGibson-SN and Isaac Sim are they have a limited number of scenes and human models. They also lack a measurement for crowdedness, which is an important factor in crowd-aware navigation problems~
\cite{karamouzas2014universal}. With that inspiration, we propose a human dynamics model that is carefully designed and included in Habitat 2.0~\cite{szot2021habitat} to create a standard and diverse crowd-aware visual navigation benchmark that closely reflects real-world conditions. Table~\ref{tab:simulator_comparison} indicates the main differences between our HabiCrowd, iGibson Social Navigation (iGibson-SN), and Isaac Sim simulator. The comparisons underscore the substantial advantages of our simulator, which offers a significantly greater number of scenes and a wider range of navigation settings, environments, and human models. %The comparisons on many factors between our HabiCrowd and iGibson Social Navigation challenge (iGibson-SN) are shown in Table~\ref{tab:simulator_comparison}.

\begin{table}[!ht]
\centering
\renewcommand\tabcolsep{1.5pt}
% \vspace{1mm}

\caption{Simulator Comparison.}
\vspace{1ex}
\resizebox{\linewidth}{!}{
\begin{tabular}{@{}lcccccc@{}}
\toprule 					&  
  \makecell{Dynamics \\ model} & \makecell{Navigation \\ settings} & \makecell{Num. \\ scenes} & 
  \makecell{Environment \\ type} & \makecell{Num. \\ humans} %& \makecell{Rendering \\ speed} 
  \\
\midrule
iGibson-SN~\cite{2022igibsonchallenge} & ORCA~\cite{van2011reciprocal} & Point-goal & 15 & residence & 3 %& up to 100~\cite{yokoyama2022benchmarking}
  \\
Isaac Sim~\cite{makoviychuk2021isaac} & \xmark & Point-goal & 7 & \makecell{office, depot\\ traffic, hospital} & 7 %& 265 [\href{https://github.com/NVIDIA-ISAAC-ROS/isaac_ros_visual_slam}{Source}] 
\\
\midrule
\makecell{\textbf{HabiCrowd} \\ (ours)} & UPL++ & \makecell{Point-goal \\ Object-goal \\ Image-goal} & 480 & \makecell{residence, gym, \\ office, studio, \\ club, restaurant} & 40 %&up to 3000~\cite{yokoyama2022benchmarking}                                     
\\
\bottomrule
\end{tabular}
}
\label{tab:simulator_comparison}
\end{table} 
\section{The HabiCrowd Simulator}
In this section, we present HabiCrowd, a crowd-aware visual navigation simulator utilizing Habitat 2.0~\cite{szot2021habitat} as the foundational simulator. The motivation behind this choice comes from the diverse environment settings and photorealistic scenes offered by Habitat 2.0~\cite{yadav2023habitat}. By integrating human dynamics into this simulator, HabiCrowd could further bridge the gap between simulation and real-life scenarios. Furthermore, Habitat 2.0 also enables parallel computing utilization~\cite{deitke2022retrospectives}, which could significantly facilitate the training procedure. We specify the configurations, trajectories, and controls of human dynamics of HabiCrowd.
%, a choice motivated by the photorealis offered by Habitat 2.0 in comparison to other simulators, as highlighted in~\cite{yokoyama2022benchmarking}. Habitat 2.0 enables parallel computing and its rendering speed can reach $1400$ frames per second~\cite{szot2021habitat}, surpasses other E-AI simulators' by a significant margin (\textit{e.g.}, $14$ times faster than iGibson~\cite{shen2021igibson}). 

\subsection{Human Dynamics Model}\label{sec:dynamics model}
\label{subsec: crowd dynamics}
We design a continuous and force-based human dynamics model, which progresses concurrently with the agent's navigation, based on the universal power law (UPL) dynamics model proposed by \cite{karamouzas2014universal}. Previously, UPL lacks the inclusion of torque components to control the facing direction of humans. To address this limitation, we further incorporate the torque dynamics model into the simulator. Consequently, our improved dynamics model is denoted as UPL++.

\begin{figure}[!ht]
    \centering
    \includegraphics[width=0.7\linewidth]{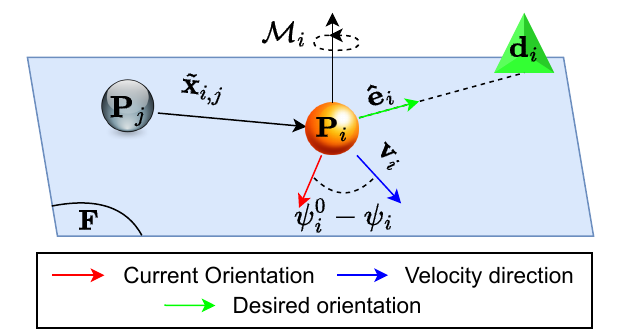}
    \vspace{2ex}
    \caption{\textbf{Human dynamics model.} We design a model to shape $\mathbf{P}_i$ towards $\mathbf{d}_i$ while avoiding collisions with $\mathbf{P}_j$. The human dynamics includes a force model and a torque model.}
    \label{fig:human_trajectory}
\end{figure}

Assume the agent is navigating in a scene $\Lambda$ with the set of obstacles $W_\Lambda$ and $n$ humans. Table~\ref{tab:notation} explains the notations in this section. We represent each human by $\mathbf{P}_i = \left (\mathbf{x}_i, \mathbf{v}_i, \hat{\mathbf{e}}_i, v_i, \psi_i, \psi_i^0, \omega_i, \omega_i^0, r_i, m_i \right )$ and consider their movements as particle kinematics on the 2D floor $\mathbf{F}$ of $\Lambda$. Each $\mathbf{P}_i$ will try to navigate to a destination $\mathbf{d}_i$. The model of $\mathbf{P}_i$ is depicted in Figure~\ref{fig:human_trajectory}. We denote $\mathbf{f}_i$ and $\mathcal{M}_i$ as the applied force and torque to each $\mathbf{P}_i$, with the subscripts "adj," "soc," and "con" corresponding "adjective," "social," and "contact", respectively.

\textbf{Force Model.} We use a force model for human entities to guide them to their destinations while minimizing collisions. The force model of $\mathbf{P}_i$ projecting on $\mathbf{F}$ can be written as:
\begin{equation}
    \mathbf{f}_i = \mathbf{f}_i^{\text{adj}} + \sum_{j=1, j\neq i}^{n} \left ( \mathbf{f}_{i,j}^{\text{soc}} + \mathbf{f}_{i,j}^{\text{con}} \right ) + \sum_{w\in W_\Lambda}\mathbf{f}_{i, w}^{\text{con}} + \zeta_i
    \label{eq:total_force}.
\end{equation}

The fluctuation force $\zeta_i=\zeta_i^\prime\vec{n}\left (\varphi_i \right )$ is included to break the symmetry of all humans, where $\zeta_i^\prime\sim\mathcal{N} (0, \sigma_{\zeta}^2 )$, $\varphi_i\sim\mathcal{U}(0, 2\pi)$, and $\vec{n} \left (\varphi_i\right )$ is the unit vector rotated by $\varphi_i$ about the origin of $\mathbf{F}$.

\begin{table}[!ht]
  \vspace{1pt}
    \centering
    % \captionsetup{type=table}
    \caption{Human Dynamics Notations.}
    \vspace{1ex}
    \resizebox{\linewidth}{!}{
    \begin{tabular}{ll|ll}
\toprule 
Not. & Description & Not. & Description \\
\midrule
$r_i$ & Particle's radius &
$\hat{\mathbf{e}}_i$ & Desired direction \\
$\mathbf{x}_i$ & Human coordinate & $\mathbf{v}_i$ & Linear velocity \\
$v_i$ & Desired velocity & $m_i$ & Human mass \\
$\omega_i$ & Current angular velocity & $\omega_i^0$ & Desired angular velocity\\
$\psi_i$ & Angular coordinate of the \\
& current orientation & $\psi_i^0$ & Angular coordinate of the \\
& desired orientation \\
\bottomrule
\end{tabular}
}
% \vspace{1ex}
\label{tab:notation}
\end{table}

\begin{figure*}[!ht]
\begin{minipage}[b]{0.60\textwidth}
\centering
\vspace{0pt}
\includegraphics[width=1.0\linewidth]{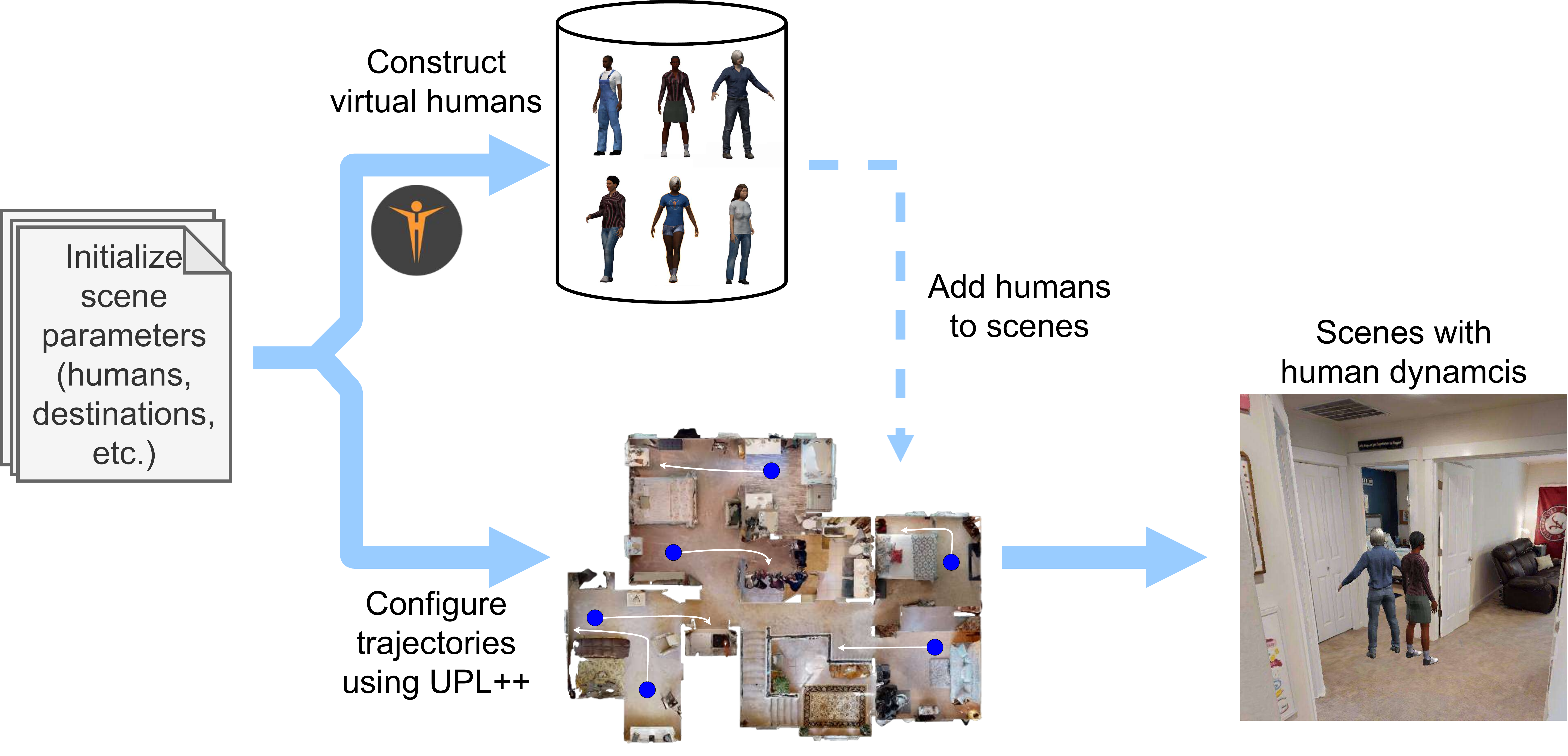}
\vspace{0.5ex}
\caption{\textbf{Dataset construction pipeline.}}
\label{fig:construction pipeline}
\end{minipage}%
\hfill
\begin{minipage}[b]{0.30\textwidth}
\centering
\vspace{0pt}
\includegraphics[width=1.0\linewidth]{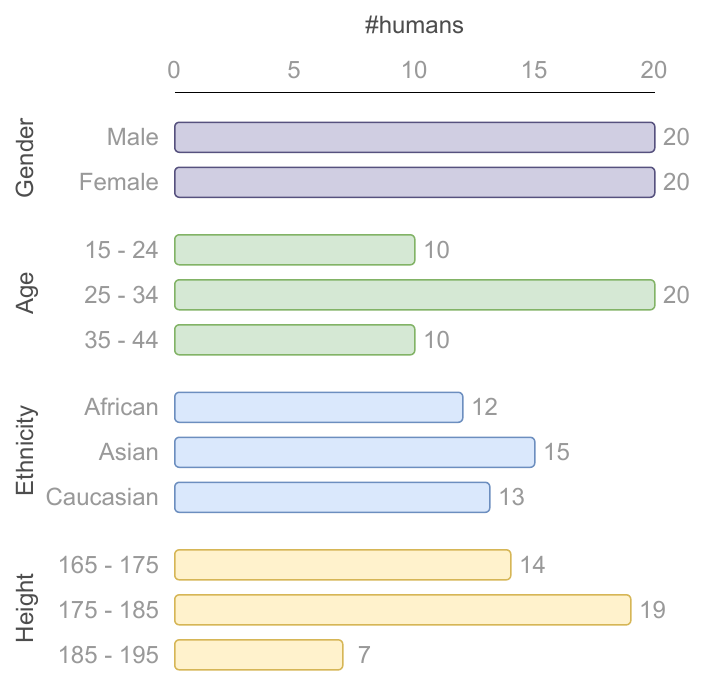}
\vspace{0.5ex}
\caption{\textbf{Human statistics.}}
\label{fig:human stats}
\end{minipage}
% \vspace{1ex}
\end{figure*}

The contact force $\mathbf{f}_{i, *}^{\text{con}}$, which is a consequence of a collision between other entities like objects/humans, indicates the corresponding physical reaction. In the former notation, $*$ is a placeholder indicating either an obstacle $w$ or another human index $j$. We rely on the core physic engine of Habitat 2.0, PyBullet~\cite{coumans2019} to model the contact force.

The adjust force $\mathbf{f}_{i}^{\text{adj}}$ shapes the agent to navigate towards its destination $\mathbf{d}_l$. Specifically, $\mathbf{f}_{i}^{\text{adj}}={m_i}/{\tau^{\text{adj}}}\left (v_i\hat{\mathbf{e}}_i-\mathbf{v}_i\right )$, where $\tau^{\text{adj}}$ is the characteristic time of adjusting force. 

Finally, we establish the social force model between $\mathbf{P}_i$ and $\mathbf{P}_j$ to prolong their time-to-collision based on their linearly-estimated positions. Let $\tau$ be the time-to-collision between two entities and $\tilde{\mathbf{x}}, \tilde{\mathbf{v}}$ be their relative position and velocity, respectively. Inspired by~\cite{karamouzas2014universal}, we define the interaction energy between $\mathbf{P}_i$ and $\mathbf{P}_j$, as follows:
\begin{equation}
    E(\tau) = \frac{k^{\text{soc}}}{\tau^2} \exp \left( -\frac{\tau}{\tau^{\text{soc}}} \right),
\label{eq:exp_tau}
\end{equation}
where $k^{\text{soc}}$ is a value for adjusting social force, $\tau^{\text{soc}}$ is the interaction time horizon. The skin-to-skin distance $h$ between two human entities after $\tau$ is given by $h(\tau) = \|\tilde{\mathbf{x}} + \tau\tilde{\mathbf{v}}\|_2-\left (r_i+r_j\right )$.
% \begin{equation}
%     h(\tau) = \|\tilde{\mathbf{x}} + \tau\tilde{\mathbf{v}}\|_2-\left (r_i+r_j\right ).
%     \label{eq:skin-to-skin}
% \end{equation}

Intuitively, the collision occurs when the skin-to-skin distance zeros; therefore, by solving the quadratic equation $h(\tau)=0$,  we obtain $\tau$ as:
\begin{equation}
    \tau\left (\tilde{\mathbf{x}}\right ) = \frac{b-\sqrt{b^2-ac}}{a},
\label{eq:tau}
\end{equation}
where $a=\|\tilde{\mathbf{v}}\|_2^2; b=-\tilde{\mathbf{x}}^\mathsf{T}\tilde{\mathbf{v}}; c=\|\tilde{\mathbf{x}}\|_2^2-(r_i+r_j)^2$. Following~\cite{karamouzas2014universal}, we calculate the desired social force as $\mathbf{f}^{\text{soc}}_{i,j}(\tilde{\mathbf{x}})  = -\nabla_{\tilde{\mathbf{x}}} E(\tau)$.

% \begin{equation}
%     \mathbf{f}^{\text{soc}}_{i,j}(\tilde{\mathbf{x}})  = -\nabla_{\tilde{\mathbf{x}}} E(\tau). 
% \label{eq:social_force}
% \end{equation}

\textbf{Torque Model.} We design a torque model to keep the orientation of $\mathbf{P}_i$ coinciding with the direction of $\mathbf{v}_i$. We denote $I_i=m_ir_i^2$ as the $\mathbf{P}_i$'s moment of inertia. Then, the torque is formulated as follows:
\begin{equation}
    \mathcal{M}_i = \mathcal{M}_i^{\text{adj}} + \sum_{j=1, j\neq i}^{n} \mathcal{M}_{i,j}^{\text{con}} + \sum_{w\in W_\Lambda} \mathcal{M}_{i,w}^{\text{con}} + \eta_i.
\label{eq:total_torque}
\end{equation}

The fluctuation torque $\eta_i\sim\mathcal{N}\left (0, \sigma_{\eta}^2\right )$ aims to break the crowd symmetry by randomly altering $\mathbf{P}_i$'s orientation.

The contact torque $\mathcal{M}_{i, *}^{\text{con}}$ is also based on the foundation simulator, which is similar to $\mathbf{f}_{i, *}^{\text{con}}$. 

The adjusting torque shapes the orientation of the human entity towards its desired orientation, which is the current linear velocity $\mathbf{v}_i$ (see Figure~\ref{fig:human_trajectory}). Additionally, we want to keep the angular velocity at a reasonable rate $\omega_i^0$. As a consequence, we design the adjusting torque as follows:
\begin{equation}
    \mathcal{M}^{\text{adj}}_i = \frac{I_i}{\tau^{\text{rot}}} \left[\left(\frac{(\psi_i^0-\psi_i)\mod 2\pi}{\pi} -1\right)\omega_i^0 - \omega_i\right],
\label{eq:adj_torque}
\end{equation}
where $\tau^{\text{rot}}$ is the characteristic time of adjusting torque.

\textbf{Computational Complexity.} It can be implied from Eqs.~\eqref{eq:total_force} and~\eqref{eq:total_torque} that the computation complexity for $\mathbf{f}_i$ and $\mathcal{M}_i$ is both $O(n)$. This is because, apart from $\mathbf{f}^{\text{soc}}_{i, j}$ and $\mathbf{f}^{\text{con}}_{i, j}$, which require $O(n)$ computations, the remaining components only need constant computations. 

We note that our UPL++ dynamics model and ORCA~\cite{van2011reciprocal} that is used in iGibson-SN~\cite{2022igibsonchallenge} share the same computational complexity of  $O(n)$. However, the major difference between our UPL++ and ORCA is the approach used to determine the dynamic terms, $\mathbf{f}_i$ and $\mathcal{M}_i$. While UPL++ directly computes these terms, ORCA employs an indirect approach by solving an LP problem. Empirical results in Section~\ref{sec: simulator} confirm that the constant factor associated with UPL++ is considerably smaller than ORCA's, indicating a computational efficiency advantage in favor of UPL++.

\subsection{Dataset Construction}
Figure~\ref{fig:construction pipeline} illustrates the pipeline to build HabiCrowd scenes. We begin with initializing parameters for each virtual human $\mathbf{P}_i$ and their destination $\mathbf{d}_i$. Next, we construct virtual humans using the MakeHuman~\cite{briceno2018makehuman}. Then the trajectories of humans are configured by using the dynamics model in Section~\ref{sec:dynamics model} and are added to the scenes of HM3D dataset~\cite{ramakrishnan2021hm3d} to create the final scenes. Note that when a human entity reaches its destination, we navigate it back to its original position and repeat this loop to maintain human dynamics during the agent's navigation.

\begin{figure*}[ht!]
  \centering
  \begin{minipage}[b]{0.345\textwidth}
    \includegraphics[width=1.0\linewidth]{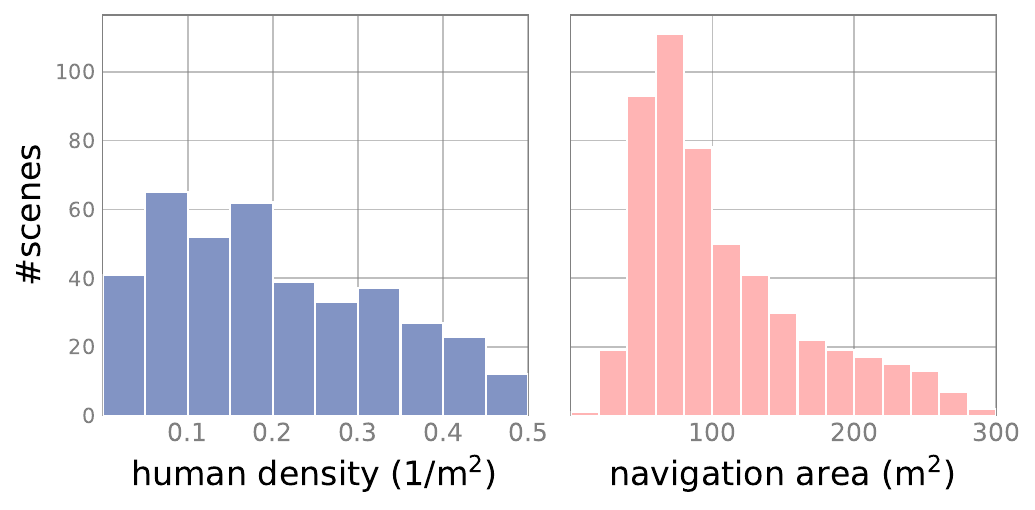}
    \vspace{0ex}
\caption{{\textbf{Navigating statistics}.}}
\label{fig:density and area nav}
  \end{minipage}
  \hfill % adds horizontal space between the images
  \begin{minipage}[b]{0.34\textwidth}
    \includegraphics[width=1.0\linewidth]{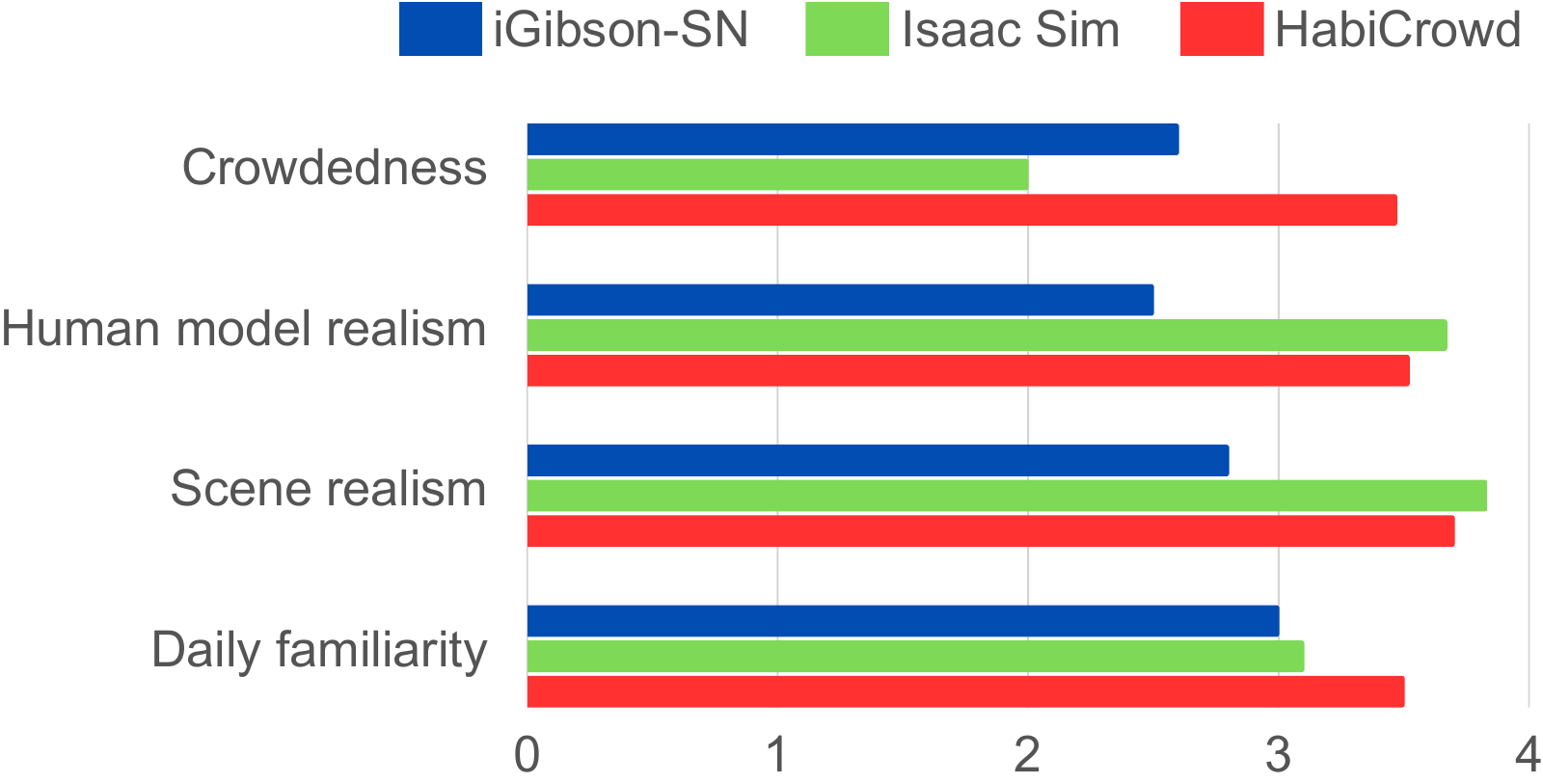}
    \vspace{1ex}
\caption{\textbf{User evaluation}.}
\label{fig:user_study}
  \end{minipage}
  \hfill % adds horizontal space between the images
  \begin{minipage}[b]{0.28\textwidth}
    \includegraphics[width=1.0\linewidth]{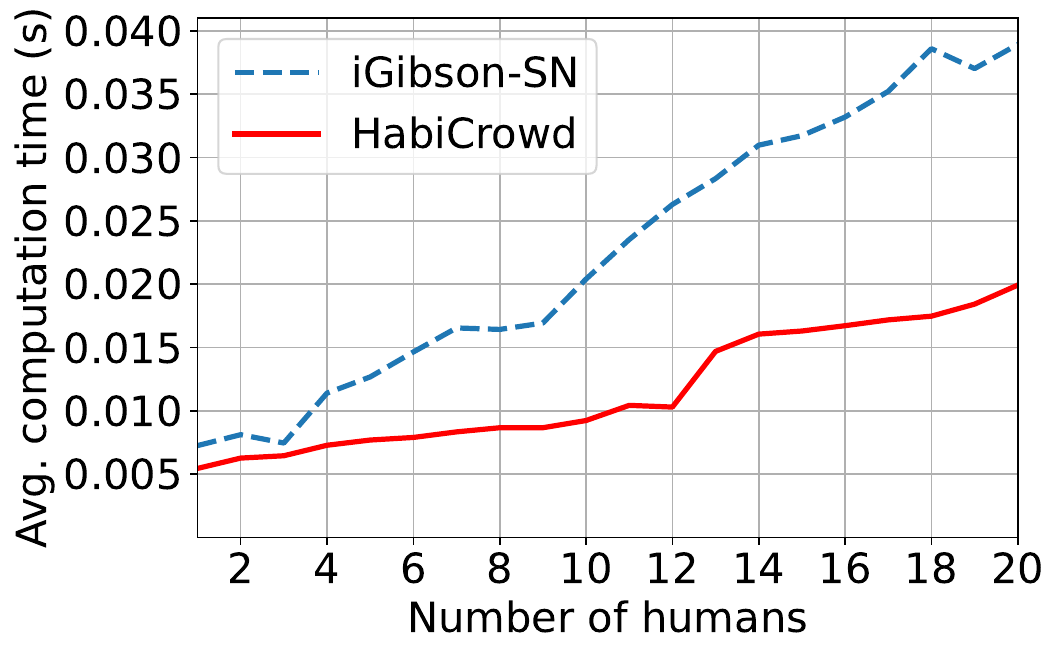}
\vspace{0.5ex}
\caption{\textbf{Computational time.}}
\label{fig:avg computational time}
  \end{minipage}
\end{figure*}

\textbf{Dataset Statistics.} We utilize $480$ scenes from HM3D~\cite{ramakrishnan2021hm3d} and $40$ human entities. The train/val/test splits are $400/40/40$. HabiCrowd's virtual humans have a gender ratio of $1/1$ and ages range from $15$ to $44$ (Figure~\ref{fig:human stats}). %In addition, we set up a variety of human poses and attire for the crowd, such as casual clothes, sports clothes, and formal suits; some examples can be seen in Figure~\ref{fig:construction pipeline} \textbf{wrong reference}. 
The human density and navigation area distributions are depicted in Figure~\ref{fig:density and area nav}. Our simulator provides a broad spectrum of navigation settings including different levels of human density, spanning from $0.1$ to $0.5$, as well as navigation areas ranging from $\SI{10}{\metre\squared}$ to $\SI{300}{\metre\squared}$. The average human density of HabiCrowd is $0.226$ humans per $\SI{}{\metre\squared}$, which is significantly greater than $0.143$, the real crowd density in the real world~\cite{deloitte2010employment}. Our simulator also has scenes with higher density (for example, $0.45$) to make the problem more challenging.

% \begin{figure}
%     \centering
%     \includegraphics[width=0.4\linewidth]{cvpr/figures/for_nips/nav_area.pdf}
%     \includegraphics[width=0.4\linewidth]{cvpr/figures/for_nips/density.pdf}
%     \caption{Distributions of navigation area and human density.}
%     \vspace{-3ex}
%     \label{fig:density and area distributions}
% \end{figure}

\section{Experiments}

\subsection{HabiCrowd Evaluation}
\label{sec: simulator}

\begin{figure}[!ht]
    \centering\includegraphics[width=0.8\linewidth]{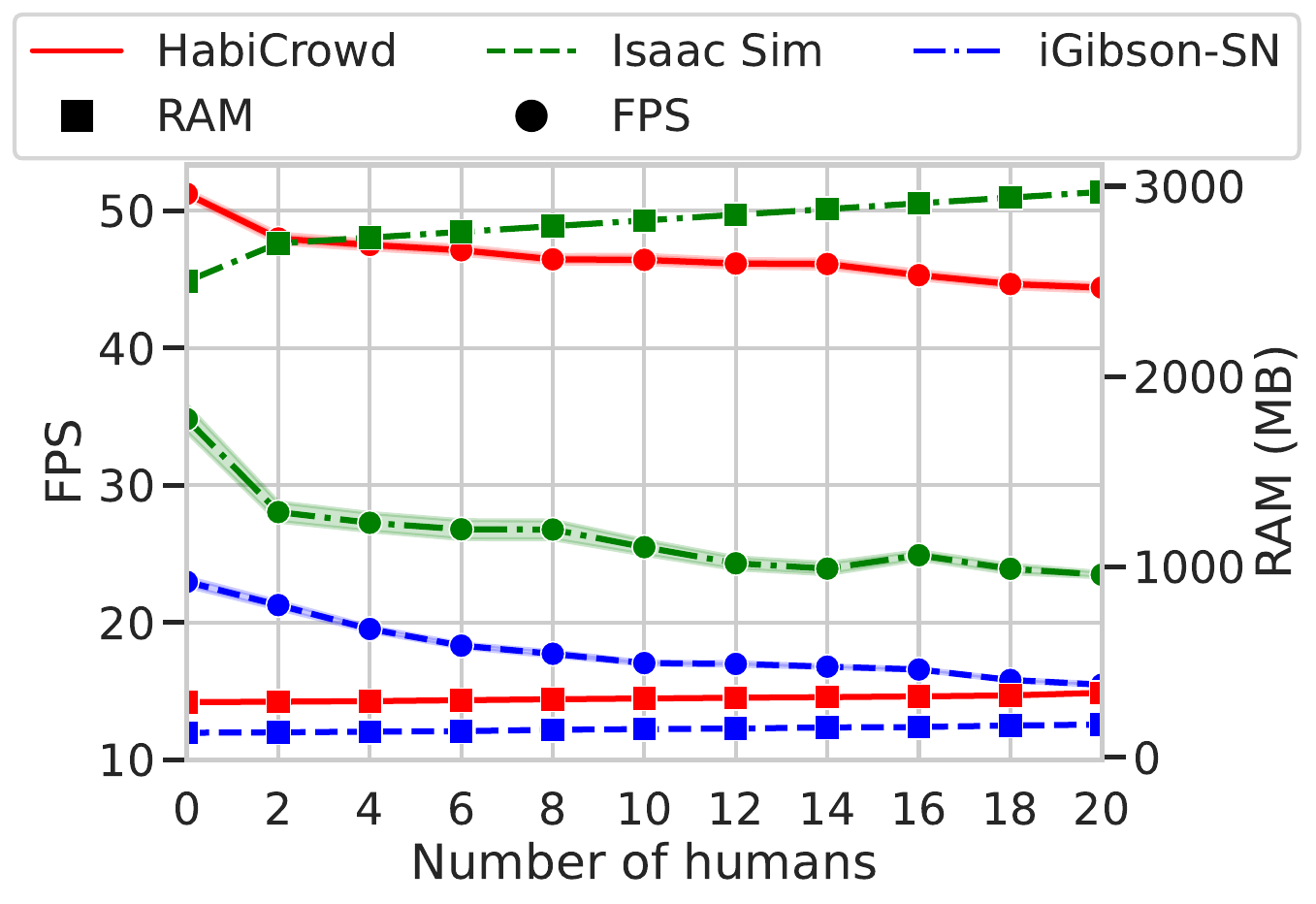}
    \vspace{1ex}
\caption{\textbf{Performance and memory benchmarks.}}
\label{fig:fps}
\end{figure}

\textbf{Performance and Memory Benchmarks.} Figure~\ref{fig:fps} compares the average rendering speed (frame per second - FPS) and memory (RAM) utilization for each scene among three simulators. The process is conducted over a set of $6$ scenes of each simulator on a cluster of $4$ RTX 3090 GPUs. We observe a downward trend in FPS and an upward trend in memory usage for all simulators as the number of virtual humans within the scenes increases. These phenomena are primarily due to the saturation of computational complexity when more humans are added to the scenes. Nevertheless, our HabiCrowd consistently exhibits an average FPS that is nearly three times higher than iGibson-SN's and two times higher than Isaac Sim's. Furthermore, in terms of memory efficiency, both our simulator and iGibson-SN stand out by requiring less than $\SI{400}{\mega\byte}$ of RAM to render each scene. In contrast, Isaac Sim necessitates over $\SI{2560}{\mega\byte}$ of RAM to render scenes with high fidelity. To conclude, our proposed simulator achieves the most efficient computational utilization compared to other baselines.

\begin{figure}[!ht]
    \centering\includegraphics[width=0.8\linewidth]{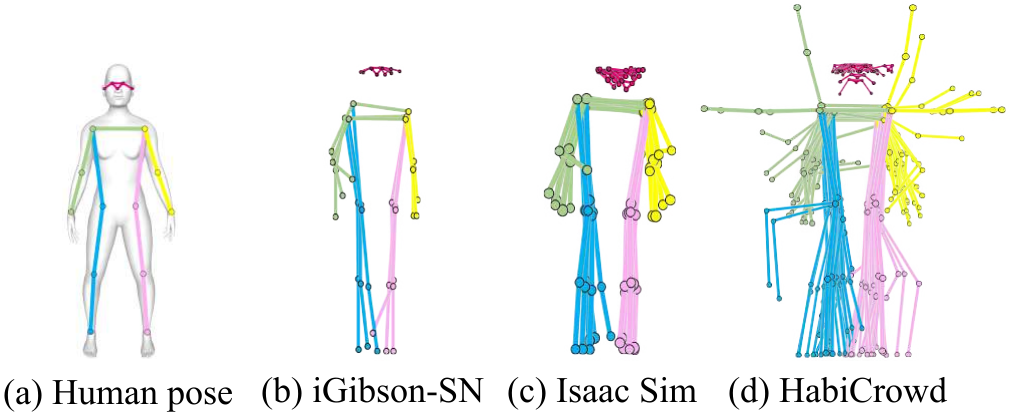}
    \vspace{1ex}\caption{\textbf{Human poses comparison.} }
    \label{fig:human_pose}
\end{figure}

%Our normalization process ensures that all human poses have the upper body facing the front (a).
\textbf{Human Dynamics Evaluation.} We leverage UMANS engine~\cite{van2020generalized} to compare our dynamic model UPL++ with ORCA dynamic model in iGibson-SN on the following metrics: \textit{i)} Mean computational time (mCT), the average time each model takes to update states for virtual humans during navigation. \textit{ii)} Collision avoidance rate (CAR), measuring how often collisions are prevented between virtual humans. \textit{iii)} Goal-reaching rate (GR), indicating the frequency of virtual humans reaching their destinations Table~\ref{tab:dynamics_evaluation} reports the evaluation results. This table shows that both iGibson-SN and our simulator achieve collision-free navigation. Although HabiCrowd demonstrates a goal-reaching rate comparable to iGibson-SN's (with a slight margin of 0.78\%), our model shows remarkable improvement in terms of computational time, which is \textit{nearly twice faster} than iGibson-SN. 

Recall from Section~\ref{sec:dynamics model} that the computational complexity of iGibson-SN and HabiCrowd is $O(n)$. However, from Figure~\ref{fig:avg computational time}, we observe that the slope of iGibson-SN is considerably steeper than HabiCrowd's, indicating that the corresponding constant factor of HabiCrowd is significantly smaller than its counterpart. As a result, the computational efficiency advantage of HabiCrowd over iGibson-SN is preserved even when the number of humans scales up.

\begin{table}[!ht]
    \centering
    \begin{minipage}{0.485\linewidth}
        \centering
        \caption{\label{tab:dynamics_evaluation}Human Dynamics Evaluation.}
        
        % \vspace{1.5ex}
        \resizebox{\linewidth}{!}{
        \begin{tabular}[t]{l@{\hskip 0.05in}c@{\hskip 0.05in}c@{\hskip 0.05in}c}
\toprule 					&  
CAR $\uparrow$ & GR $\uparrow$ & mCT $\downarrow$ \\
\midrule
iGibson-SN~\cite{2022igibsonchallenge} & \bf{100\%} & \bf{94.1\%} & 0.023 \\
HabiCrowd (our) & \bf{100\%} & 93.3\% & \bf{0.012} \\
\bottomrule
\end{tabular}}
    \end{minipage}
    \hspace{0.01\linewidth} % Space between the two minipages
    \begin{minipage}{0.485\linewidth}
        \centering
        \caption{Human Density Analysis.}
        \label{tab:ablation_density}
        % \vspace{1.5ex}
        \resizebox{\linewidth}{!}{
        \begin{tabular}[t]{@{}r@{\hskip 0.05in}c@{\hskip 0.05in}c@{\hskip 0.05in}c@{}}
            \toprule
            \diagbox{Baseline}{Density} & \textless 0.2 & 0.2-0.3 & \textgreater 0.3 \\
            \midrule
            LB-WPN~\cite{tolani2021visual} & {16.37} & {12.64} & {6.28} \\
            DOA~\cite{yokoyama2022benchmarking} & {17.78} & {13.25} & {7.36} \\
            DD-PPO~\cite{wijmans2019dd} & {15.83} & {11.79} & {7.07} \\
            DS-RNN~\cite{liu2021decentralized} & {14.95} & {7.51} & {4.24} \\
            OVRL-v2~\cite{yadav2023ovrl} & {\textbf{22.38}} & {\textbf{16.49}} & {\textbf{11.27}} \\
            \bottomrule
        \end{tabular}}
    \end{minipage}
\end{table}

\textbf{User Study.} We provide a user study with {52} participants from diverse professions, such as designers, animators, developers, and marketers. We ask them to rate HabiCrowd, Isaac Sim, and iGibson-SN on four criteria: crowdedness, human model realism, scene realism, and daily familiarity. Figure~\ref{fig:user_study} shows that in all aspects, our simulator is preferred over iGibson-SN. In addition, HabiCrowd shows comparable levels of human model realism and scene realism, with a higher familiarity compared to Isaac Sim.

\textbf{Human Poses Comparison.} We extract human poses from all datasets by using the PCT approach~\cite{geng2023human} and provide the qualitative analysis in Figure~\ref{fig:human_pose}. The findings reveal that our simulator showcases a greater degree of diversity in human poses when compared to the other simulators. Consequently, HabiCrowd captures a wider range of human pose variations, therefore, reflecting a more comprehensive representation of real-life human settings.

\begin{table}[!ht]
    \centering
    \caption{Visual Navigation Results.}
\label{tab:point nav result}
% \vspace{1.5ex}
\resizebox{\linewidth}{!}{
    \begin{tabular}[t]{@{}rcccccccc@{}}
\toprule 
& \multicolumn{4}{c}{Point Navigation} & \multicolumn{4}{c}{Object Navigation} \\
            \cmidrule(lr){2-5} \cmidrule(lr){6-9}
            
&  
CPD $\downarrow$ & SPL $\uparrow$ & SR $\uparrow$ & DTG $\downarrow$ & CPD $\downarrow$ & SPL $\uparrow$ & SR $\uparrow$ & DTG $\downarrow$\\
\midrule
% Random & \hspace{5mm} - & 0.000 & 0.000 & 7.733 \\
LB-WPN~\cite{tolani2021visual} & {0.0521} & {78.13}  & {91.84} & {0.6048} & {0.0462} & {7.989} & {15.31} & {5.801} \\
DOA~\cite{yokoyama2022benchmarking} & {0.0626} &  {80.44} & {92.96} & {0.5809} & {0.0441} & {8.278} & {16.57} & {5.742} \\
DD-PPO~\cite{wijmans2019dd} & {0.0617} & {79.34} & {92.55} & {0.5991} & {0.0415} & {7.424} & {14.93} & {5.956} \\
DS-RNN~\cite{liu2021decentralized} & {0.0476} & {76.82} & {90.73} & {0.6674} & {0.0305} & {6.078} & {12.34} & {5.972} \\
OVRL-v2~\cite{yadav2023ovrl} & {\textbf{0.0322}} & {\textbf{82.65}} & {\textbf{94.61}} & {\textbf{0.4895}} & {\textbf{0.0229}} & {\textbf{10.267}} & {\textbf{20.79}} & {\textbf{5.413}}\\
% DAViT (ours) & \textbf{0.0295} & \textbf{82.78} & \textbf{94.68} & \textbf{0.4737} \\
\bottomrule
\end{tabular}
}
% \vspace{-1.5ex}
\end{table}

\begin{table*}[!ht]
\small
\centering
\scriptsize
\renewcommand\tabcolsep{2.5pt}
\renewcommand{\arraystretch}{0.91}
\caption{Impact of Collision Penalty.}
\vspace{1.5ex}
\label{table:ablation_collision_penalty}
% \resizebox{\textwidth}{!}{%
\begin{tabular}{{r}cccc|cccccccccccc}
    \toprule
    & \multicolumn{4}{c|}{Without our reward model} & \multicolumn{12}{c}{With our reward model} \\
    \cmidrule(lr){2-5} \cmidrule(lr){6-17}
    & \multicolumn{4}{c|}{$R^{\text{c}}=0.0$} & \multicolumn{4}{c}{$R^{\text{c}}=-10^{-4}$} & \multicolumn{4}{c}{$R^{\text{c}}=-10^{-3}$} & \multicolumn{4}{c}{$R^{\text{c}}=-10^{-2}$} \\
    \cmidrule(lr){2-5} \cmidrule(lr){6-9} \cmidrule(lr){10-13} \cmidrule(lr){14-17}
    & CPD$\downarrow$ & SPL$\uparrow$ & SR$\uparrow$ & DTG$\downarrow$
    & CPD$\downarrow$ & SPL$\uparrow$ & SR$\uparrow$ & DTG$\downarrow$
    & CPD$\downarrow$ & SPL$\uparrow$ & SR$\uparrow$ & DTG$\downarrow$
    & CPD$\downarrow$ & SPL$\uparrow$ & SR$\uparrow$ & DTG$\downarrow$ \\
    \midrule
    LB-WPN~\cite{tolani2021visual} & {0.076} & {5.90} & {10.62} & {6.19} & {0.046} & {7.99} & {15.31} & {5.80} & {0.036} & {4.78} & {8.99} & {6.38} & {0.021} & {3.44} & {6.04} & {6.78}  \\
    DOA~\cite{yokoyama2022benchmarking} & {0.085} & {6.89} & {12.68} & {5.95}  & {0.044} & {8.28} & {16.57} & {5.74} & {0.028} & {5.22} & {9.35} & {6.36} & {0.031} & {3.74} & {6.29} & {6.80} \\
    DD-PPO~\cite{wijmans2019dd} & {0.059} & {6.23}  & {11.74} & {6.03} & {0.042} & {7.42} & {14.93} & {5.96} & {0.032} & {4.54} & {8.26} & {6.41} & {0.027} & {3.13} & {5.83}  &  {6.89}\\
    DS-RNN~\cite{liu2021decentralized} & {0.055} & {5.28} & {9.26} & {6.14} & {0.031} & {6.08} & {12.34} & {5.97} & {0.026} & {3.48} & {7.42} & {6.67} & {0.018} & {2.15} & {4.73} & {6.94} \\
    % E2E-PPO~\cite{ma2019using} & 0.0411 & 4.108 & 6.871 & 6.500 & 0.0366 & 2.297 & 2.732 & 7.130 & 0.3537 & 1.214 & 1.325 & 7.395 \\
    OVRL-v2~\cite{yadav2023ovrl} & {\bf{0.041}} & {\bf{7.57}} & {\bf{14.09}} & {\bf{5.86}} & {\bf{0.023}} & {\bf{10.27}} & {\bf{20.79}} & {\bf{5.41}} & {\bf{0.020}} & {\bf{6.15}} & {\bf{11.73}} & {\bf{6.27}} & {\bf{0.014}} & {\bf{4.22}} & {\bf{7.79}} & {\bf{6.68}} \\
    \bottomrule	
\end{tabular}
% }
% \vspace{1ex}
% \vspace{-5ex}
\end{table*}

\vspace{-1.1ex}
\subsection{Crowd-aware Visual Navigation}
\vspace{-1.1ex}

\textbf{Experimental Settings.} %We examine three crowd-aware visual navigation tasks upon HabiCrowd, namely, \textit{point navigation}~\cite{wijmans2019dd}, \textit{object navigation}~\cite{chaplot2020object}, \textit{instance image-goal navigation}~\cite{krantz2022instance}. 
We examine two crowd-aware visual navigation tasks upon HabiCrowd, namely, \textit{point navigation}~\cite{wijmans2019dd} and \textit{object navigation}~\cite{chaplot2020object}. The agent is initialized at a pre-defined location in a human-crowded scene and asked to navigate to an objective. At each timestep, the agent receives RGB and Depth observations. We introduce a new objective that requires the agent to minimize the number of collisions with virtual humans. %Each collision is counted when the skin-to-skin distance between the agent and any human zeros. %The objective varies depending on the task, wherein it corresponds to an $xyz$-coordinate for point navigation and a specified object for object navigation. 

%that depict its egocentric viewpoint. 

Denote $d_{t}$ as the distance from the agent to the goal at timestep $t$, we design the reward model as: $    r_t = R^{\text{c}}\mathbb{I}^{\text{c}}_t + R^{\text{s}}\mathbb{I}^{\text{s}}_t + R^{\text{f}}\times(d_{t}-d_{t-1})
$, where $R^{\text{c}}$ is the collision penalty, $R^{\text{s}}$ is the reward for completing the task, $R^{\text{f}}$ is the shaping reward for moving closer to the objective, and $\mathbb{I}^{\text{c}}_t$, $\mathbb{I}^{\text{s}}_t$ are the indicators of collision and success of the current step, respectively. This reward model helps balance between navigating towards the objective and avoiding collisions. We train all methods using the above reward model. 
Followed by~\cite{chaplot2020object}, we set $R^{\text{s}}=2.5, R^{\text{f}}=1.0$. The default value of $R^{\text{c}}$ is set to $-10^{-4}$. For the human dynamics, we set $\tau^{\text{adj}}=0.5, \tau^{\text{soc}}=3.0, \tau^{\text{rot}}=0.2$ and $k^{\text{soc}}=1.5$ in all setups. Each method is trained in an end-to-end fashion using a total of 30M frames. We run each experiment on a computing cluster utilizing a node of four 48GB-RAM V100 GPUs.

\textbf{Baselines.}
We compare the following baselines:
\textit{i)} DS-RNN~\cite{liu2021decentralized}: A spatial and temporal learning-based agent%, which captures the human-robot relationships. The agent then employs RNN models to infer the next action, which is trained with PPO algorithm~\cite{schulman2017proximal}
. \textit{ii)} DD-PPO~\cite{wijmans2019dd}: A widely-used state-of-the-art optimization algorithm in E-AI tasks. %The algorithm is based on distributed computing to accelerate policy learning.
 % 3) E2E-PPO~\cite{ma2019using}: An approach to tackle the crowd-aware navigation task using egocentric inputs. Using PPO as the optimizer, the agent extracts visual features via a variational autoencoder~\cite{kingma2013auto}.
\textit{iii)} LB-WPN~\cite{tolani2021visual}: A method that combines a learning-based perception module and a model-based planning module for autonomous navigation among humans. \textit{iv)} DOA~\cite{yokoyama2022benchmarking}: The winning entry of the 2021 iGibson-SN challenge~\cite{2022igibsonchallenge} that applies the dynamic obstacle augmentation method. \textit{v)} OVRL-v2~\cite{yadav2023ovrl}: A recent baseline utilizing the vision transformer approach (ViT).

\textbf{Metrics.} 
We employ three metrics in traditional visual navigation tasks~\cite{chaplot2020object}, namely, \textit{i)} Distance to goal (DTG), \textit{ii)} Success rate (SR - in \%), \textit{iii)} Success path length~\cite{anderson2018evaluation} (SPL - in \%). Additionally, we introduce \textit{iv)} \textbf{\textit{Collisions per distance (CPD)}}, a new metric to quantify collisions occurring between the navigating agent and virtual humans. This metric is computed by dividing the number of collisions between the agent and humans by the distance traveled by the agent.
%Distance to goal~\cite{chaplot2020object} (DTG): measures the distance ($\SI{}{\metre}$) between the agent and the goal at the end of the episode; 3) Success path length~\cite{anderson2018evaluation} (SPL): scales success rate (\%) with the number of steps in an episode, which indicate whether successful waypoints are optimal or not; and 4) Collisions per distance (CPD): is measured by the number of times the agent collides with humans divided by its distance traveled ($1/\SI{}{\metre}$).

\textbf{Point Navigation Results.} We present the point navigation results in Table~\ref{tab:point nav result}. Overall, every baseline achieves decent performance, %implying that point navigation is well-addressed. 
OVRL-v2 achieves the most comprehensive performance in terms of all metrics. %Notably, OVRL-v2 achieves the best CPD, which improves at least 44\% over other methods. We explain this phenomenon as follows. 
Since HabiCrowd does not provide dynamics information regarding virtual humans to the agent due to our aim of enabling more realistic navigation settings, baselines such as DS-RNN~\cite{liu2021decentralized}, LB-WPN~\cite{tolani2021visual} are unable utilize human dynamics models to address the task effectively. On the other hand, OVRL-v2 leverages ViT, a robust vision backbone capable of extracting entities (such as virtual humans) from visual inputs as additional clues, thereby proficiently handling the task.

\textbf{Object Navigation Results.}
Table~\ref{tab:point nav result} illustrates the results on the HabiCrowd dataset by all approaches. It is observable that object navigation poses a significantly greater challenge than point navigation, as evidenced by the decline in evaluation metrics for all baseline methods. The decrease in performance is aligned with the expectations outlined in~\cite{khandelwal2022simple}. The results show that OVRL-v2 has the best success rate, outperforming the second-best algorithm (DOA) by {$4.22\%$}. %, DS-RNN, and DD-PPO. %OVRL-v2 agent also has the lowest DTG ($\SI{0.098}{\metre}$ less than the runner-up DOA), the lowest CPD ($0.0017$ less than the runner-up DS-RNN), the highest SPL ($2.193\%$ higher than DD-PPO). 
%We see that methods utilizing human dynamics to solve the task (DS-RNN, LB-WPN, DOA, and OVRL-v2) achieve higher success rates than the other (DD-PPO), indicating the importance of learning human dynamics.

% {\textbf{Instance Image Navigation Results.} We present instance image-goal navigation results in Table~\ref{tab: nav result}. The outcomes align with the recent Habitat challenge~\cite{habitatchallenge2023}, in which the best method only achieves a success rate of about 6\%. Consequently, the image-goal navigation exhibits unsatisfactory performance, suggesting a need for improvements in both HabiCrowd and HM3D.}

\textbf{How does human density affect crowd navigation?}
Table~\ref{tab:ablation_density} shows the success rates when we test all methods with different human density setups. OVRL-v2 stably outperforms other algorithms by {$4.60\%$, $3.24\%$, and $3.91\%$} when the human density are from $<0.2$; $0.2 - 0.3$; and $>0.3$, respectively. In general, we observe that the higher the human density, the more challenging our task becomes as the success rates of all methods decrease. %In general, OVRL-v2 remains stable and outperforms other baselines.

% \begin{figure}[t]
%     \centering
%     \includegraphics[width=0.9\linewidth]{cvpr/figures/for_nips/attention.pdf}
%     \vspace{1.5ex}
%    \caption{\textbf{Human attention.} We depict the attention mechanism used by ViT backbone of OVRL-v2.
%    }
%    \label{fig:patch-attention}
% \end{figure}

\textbf{Does our reward model help?} Table~\ref{table:ablation_collision_penalty} reports the results with and without our reward model for all baselines. Clearly, our reward model improves CPD, SPL, and SR performance metrics across all methods. We explain as follows. The introduced quantity $R^{\text{c}}$ contributes to agents' regularization towards avoiding collisions with virtual humans, consequently improving performances in unseen cases.

\textbf{How does collision penalty affect crowd navigation?} We study the impact of $R^{\text{c}}$ on our navigation problem. %Table~\ref{table:ablation_collision_penalty} demonstrates the performances of all baselines when decreasing $R^{\text{c}}$ from $-10^{-4}$ to $-10^{-2}$. 
We observe in Table~\ref{table:ablation_collision_penalty} that starting from $R^{\text{c}}=-10^{-4}$ to $-10^{-2}$, all baselines experience a downward tendency in success rates and an upward tendency in collision avoidance. This observation is to be expected, considering that $R^{\text{c}}$ serves as a regularizing factor for the collision avoidance component. Consequently, as $R^{\text{c}}$ diminishes, the focus of the baselines shifts towards prioritizing collision avoidance with virtual humans rather than navigating towards the goal. %Across all scenarios examined, OVRL-v2 consistently exhibits the most comprehensive performance across all evaluation metrics.

% \textbf{Exp with and without using our reward function?}

% \textbf{Will attention mechanism help?} Figure~\ref{fig:patch-attention} demonstrates how the attention mechanism works in ViT backbone of OVRL-v2. We can see that the agent trained with the transformer-based method learns to attend to virtual humans of the scene to extract more useful information to complete the navigation tasks. However, it is noticeable that ViT also allocates attention to image patches representing the wall and ceiling, indicating potential areas for improvements of the attention mechanism.

\textbf{What is the drawback of current methods?} A notable weakness of all current baselines is that they do not possess an \textit{explicit} strategy to determine human dynamics from RGB-D egocentric observations. Although OVRL-v2 leverages a robust vision backbone to extract human dynamics, the employed mechanism remains \textit{implicit} and requires refinement.

\section{Discussion}
From the experiments, we can see that the point-goal crowd-aware navigation task is well-addressed, while the object-goal task remains a challenge. % in which we can observe that there is still a large room for improving navigating metrics as well as collision avoidance. 
We have highlighted the critical issue in 3D crowd-aware visual navigation tasks, emphasizing the necessity of explicitly extracting human dynamics solely from RGB-D observations. This direction should be studied as making efforts towards solving crowd-aware navigation in real-life settings, where robots can only perceive human behaviors via sensor inputs without the assumption of a known human dynamics model.

Training navigation agents in simulations like Habitat shows promise, yet transferring these learned policies to real-world environments, such as on robotic platforms, continues to be a significant challenge~\cite{ramakrishnan2022poni}. Early-stage sim2real initiatives, including ROS interfaces, confront issues like physics discrepancies~\cite{chen2022ros}. Furthermore, the gap between simulated and practical environments is widened by differences in visual appearance, absence of real-world noise, and oversimplified physical interactions, leading to poor real-world generalization of simulation-learned navigation policies~\cite{rosano2021embodied}. In line with these efforts, our work aims to mitigate these issues by incorporating human dynamics into the Habitat simulator to create a more realistic bridge between the simulated world and the practical environment. We believe integrating human dynamics into HabiCrowd represents an encouraging step towards developing robots capable of learning in dynamic, human-aware environments that more closely replicate the natural environment encountered daily.
\section{Conclusion and Future Work}
We have presented HabiCrowd, a new crowd-aware simulator built upon Habitat 2.0. HabiCrowd has a robust human dynamic model and a diverse range of virtual humans compared to related simulators. The intensive experimental results indicate that our simulator utilizes computational resources more efficiently than its counterparts. We evaluate two visual navigation tasks and show that the recognition of virtual humans depicted in egocentric inputs is essential. Nevertheless, HabiCrowd still has limitations. While humans exhibit natural movements in daily lives, the constraints imposed by Habitat 2.0~\cite{szot2021habitat} only allow for rigid movements. This assumption falls considerably behind the complexity of real-life practice and thus requires further enhancements. %Finally, applying the learned policy from Habicrowd to the real world remains an open problem for future work.

 %Finally, HabiCrowd facilitates studies pertaining to human density, human recognition, and collision avoidance, thus enabling further investigations into these crucial aspects of human-aware E-AI systems. %Although it is acceptable to regard rigid entities as humanoids~\cite{bennewitz2005learning} that robots can learn to perceive, 

\bibliographystyle{class/IEEEtran}
\bibliography{class/IEEEabrv,class/reference}

\begin{thebibliography}{10}
\providecommand{\url}[1]{#1}
\csname url@rmstyle\endcsname
\providecommand{\newblock}{\relax}
\providecommand{\bibinfo}[2]{#2}
\providecommand\BIBentrySTDinterwordspacing{\spaceskip=0pt\relax}
\providecommand\BIBentryALTinterwordstretchfactor{4}
\providecommand\BIBentryALTinterwordspacing{\spaceskip=\fontdimen2\font plus
\BIBentryALTinterwordstretchfactor\fontdimen3\font minus \fontdimen4\font\relax}
\providecommand\BIBforeignlanguage[2]{{%
\expandafter\ifx\csname l@#1\endcsname\relax
\typeout{** WARNING: IEEEtran.bst: No hyphenation pattern has been}%
\typeout{** loaded for the language `#1'. Using the pattern for}%
\typeout{** the default language instead.}%
\else
\language=\csname l@#1\endcsname
\fi
#2}}

\bibitem{yadav2023ovrl}
K.~Yadav, A.~Majumdar, R.~Ramrakhya, N.~Yokoyama, A.~Baevski, Z.~Kira, O.~Maksymets, and D.~Batra, ``Ovrl-v2: A simple state-of-art baseline for imagenav and objectnav,'' \emph{arXiv preprint arXiv:2303.07798}, 2023.

\bibitem{srivastava2022behavior}
S.~Srivastava, C.~Li, M.~Lingelbach, R.~Mart{\'\i}n-Mart{\'\i}n, F.~Xia, K.~E. Vainio, Z.~Lian, C.~Gokmen, S.~Buch, K.~Liu, \emph{et~al.}, ``Behavior: Benchmark for everyday household activities in virtual, interactive, and ecological environments,'' in \emph{CoRL}, 2022.

\bibitem{yadav2023habitat}
K.~Yadav, R.~Ramrakhya, S.~K. Ramakrishnan, T.~Gervet, J.~Turner, A.~Gokaslan, N.~Maestre, A.~X. Chang, D.~Batra, M.~Savva, \emph{et~al.}, ``Habitat-matterport 3d semantics dataset,'' in \emph{CVPR}, 2023.

\bibitem{wijmans2019dd}
E.~Wijmans, A.~Kadian, A.~Morcos, S.~Lee, I.~Essa, D.~Parikh, M.~Savva, and D.~Batra, ``{DD-PPO:} learning near-perfect pointgoal navigators from 2.5 billion frames,'' in \emph{ICLR}, 2020.

\bibitem{chaplot2020object}
D.~S. Chaplot, D.~P. Gandhi, A.~Gupta, and R.~R. Salakhutdinov, ``Object goal navigation using goal-oriented semantic exploration,'' \emph{NeurIPS}, 2020.

\bibitem{gupta2017cognitive}
S.~Gupta, J.~Davidson, S.~Levine, R.~Sukthankar, and J.~Malik, ``Cognitive mapping and planning for visual navigation,'' in \emph{CVPR}, 2017.

\bibitem{ramakrishnan2021hm3d}
S.~K. Ramakrishnan, A.~Gokaslan, E.~Wijmans, O.~Maksymets, A.~Clegg, J.~Turner, E.~Undersander, W.~Galuba, A.~Westbury, A.~X. Chang, \emph{et~al.}, ``Habitat-matterport 3d dataset (hm3d): 1000 large-scale 3d environments for embodied ai,'' in \emph{NeurIPS Datasets and Benchmarks Track}, 2021.

\bibitem{bansal2020combining}
S.~Bansal, V.~Tolani, S.~Gupta, J.~Malik, and C.~Tomlin, ``Combining optimal control and learning for visual navigation in novel environments,'' in \emph{CoRL}, 2020.

\bibitem{liu2021decentralized}
S.~Liu, P.~Chang, W.~Liang, N.~Chakraborty, and K.~Driggs-Campbell, ``Decentralized structural-rnn for robot crowd navigation with deep reinforcement learning,'' in \emph{ICRA}, 2021.

\bibitem{monaci2022dipcan}
G.~Monaci, M.~Aractingi, and T.~Silander, ``Dipcan: Distilling privileged information for crowd-aware navigation,'' \emph{RSS}, 2022.

\bibitem{bonin2008visual}
F.~Bonin-Font, A.~Ortiz, and G.~Oliver, ``Visual navigation for mobile robots: A survey,'' \emph{Journal of Intelligent and Robotic Systems}, 2008.

\bibitem{chen2019crowd}
C.~Chen, Y.~Liu, S.~Kreiss, and A.~Alahi, ``Crowd-robot interaction: Crowd-aware robot navigation with attention-based deep reinforcement learning,'' in \emph{ICRA}, 2019.

\bibitem{kretzschmar2016socially}
H.~Kretzschmar, M.~Spies, C.~Sprunk, and W.~Burgard, ``Socially compliant mobile robot navigation via inverse reinforcement learning,'' \emph{IJRR}, 2016.

\bibitem{dugas2022navdreams}
D.~Dugas, O.~Andersson, R.~Siegwart, and J.~J. Chung, ``Navdreams: Towards camera-only rl navigation among humans,'' \emph{arXiv preprint arXiv:2203.12299}, 2022.

\bibitem{alahi2016social}
A.~Alahi, K.~Goel, V.~Ramanathan, A.~Robicquet, L.~Fei-Fei, and S.~Savarese, ``Social lstm: Human trajectory prediction in crowded spaces,'' in \emph{CVPR}, 2016.

\bibitem{chen2017socially}
Y.~F. Chen, M.~Everett, M.~Liu, and J.~P. How, ``Socially aware motion planning with deep reinforcement learning,'' in \emph{IROS}, 2017.

\bibitem{2022igibsonchallenge}
\BIBentryALTinterwordspacing
C.~Li, J.~Jang, F.~Xia, R.~Martín-Martín, C.~D'Arpino, A.~Toshev, A.~Francis, E.~Lee, and S.~Savarese, ``{iGibson Challenge 2022},'' 2022. [Online]. Available: \url{http://svl.stanford.edu/igibson/challenge.html}
\BIBentrySTDinterwordspacing

\bibitem{makoviychuk2021isaac}
V.~Makoviychuk, L.~Wawrzyniak, Y.~Guo, M.~Lu, K.~Storey, M.~Macklin, D.~Hoeller, N.~Rudin, A.~Allshire, A.~Handa, \emph{et~al.}, ``Isaac gym: High performance gpu-based physics simulation for robot learning,'' \emph{arXiv preprint arXiv:2108.10470}, 2021.

\bibitem{yokoyama2022benchmarking}
N.~Yokoyama and et.al., ``Benchmarking augmentation methods for learning robust navigation agents: the winning entry of the 2021 igibson challenge,'' in \emph{IROS}, 2022.

\bibitem{hirose2019deep}
N.~Hirose, F.~Xia, R.~Mart{\'\i}n-Mart{\'\i}n, A.~Sadeghian, and S.~Savarese, ``Deep visual mpc-policy learning for navigation,'' \emph{RA-L}, 2019.

\bibitem{li2015vision}
Z.~Li, C.~Yang, C.-Y. Su, J.~Deng, and W.~Zhang, ``Vision-based model predictive control for steering of a nonholonomic mobile robot,'' \emph{IEEE Transactions on Control Systems Technology}, 2015.

\bibitem{lucia2020stability}
S.~Lucia, S.~Subramanian, D.~Limon, and S.~Engell, ``Stability properties of multi-stage nonlinear model predictive control,'' \emph{Systems \& Control Letters}, 2020.

\bibitem{mnih2015human}
V.~Mnih, K.~Kavukcuoglu, D.~Silver, A.~A. Rusu, J.~Veness, M.~G. Bellemare, A.~Graves, M.~Riedmiller, A.~K. Fidjeland, G.~Ostrovski, \emph{et~al.}, ``Human-level control through deep reinforcement learning,'' \emph{Nature}, 2015.

\bibitem{yang2018visual}
W.~Yang, X.~Wang, A.~Farhadi, A.~Gupta, and R.~Mottaghi, ``Visual semantic navigation using scene priors,'' in \emph{ICLR}, 2019.

\bibitem{khandelwal2022simple}
A.~Khandelwal, L.~Weihs, R.~Mottaghi, and A.~Kembhavi, ``Simple but effective: Clip embeddings for embodied ai,'' in \emph{CVPR}, 2022.

\bibitem{ramrakhya2023pirlnav}
R.~Ramrakhya, D.~Batra, E.~Wijmans, and A.~Das, ``Pirlnav: Pretraining with imitation and rl finetuning for objectnav,'' \emph{arXiv preprint arXiv:2301.07302}, 2023.

\bibitem{ferrer2013robot}
G.~Ferrer, A.~Garrell, and A.~Sanfeliu, ``Robot companion: A social-force based approach with human awareness-navigation in crowded environments,'' in \emph{IROS}, 2013.

\bibitem{trautman2010unfreezing}
P.~Trautman and A.~Krause, ``Unfreezing the robot: Navigation in dense, interacting crowds,'' in \emph{IROS}, 2010.

\bibitem{trautman2020real}
P.~Trautman and K.~Patel, ``Real time crowd navigation from first principles of probability theory,'' in \emph{ICAPS}, 2020.

\bibitem{duan2022survey}
J.~Duan, S.~Yu, H.~L. Tan, H.~Zhu, and C.~Tan, ``A survey of embodied ai: From simulators to research tasks,'' \emph{IEEE Transactions on Emerging Topics in Computational Intelligence}, 2022.

\bibitem{koenig2004design}
N.~Koenig and A.~Howard, ``Design and use paradigms for gazebo, an open-source multi-robot simulator,'' in \emph{IROS}, 2004.

\bibitem{qiu2017unrealcv}
W.~Qiu, F.~Zhong, Y.~Zhang, S.~Qiao, Z.~Xiao, T.~S. Kim, and Y.~Wang, ``Unrealcv: Virtual worlds for computer vision,'' in \emph{ACMMM}, 2017.

\bibitem{kolve2017ai2}
E.~Kolve, R.~Mottaghi, W.~Han, E.~VanderBilt, L.~Weihs, A.~Herrasti, D.~Gordon, Y.~Zhu, A.~Gupta, and A.~Farhadi, ``Ai2-thor: An interactive 3d environment for visual ai,'' \emph{arXiv preprint arXiv:1712.05474}, 2017.

\bibitem{xia2018gibson}
F.~Xia, A.~R. Zamir, Z.~He, A.~Sax, J.~Malik, and S.~Savarese, ``Gibson env: Real-world perception for embodied agents,'' in \emph{CVPR}, 2018.

\bibitem{szot2021habitat}
A.~Szot, A.~Clegg, E.~Undersander, E.~Wijmans, Y.~Zhao, J.~Turner, N.~Maestre, M.~Mukadam, D.~S. Chaplot, O.~Maksymets, \emph{et~al.}, ``Habitat 2.0: Training home assistants to rearrange their habitat,'' \emph{NeurIPS}, 2021.

\bibitem{karamouzas2014universal}
I.~Karamouzas, B.~Skinner, and S.~J. Guy, ``Universal power law governing pedestrian interactions,'' \emph{Physical Review Letters}, 2014.

\bibitem{van2011reciprocal}
J.~Van Den~Berg and et.al., ``Reciprocal n-body collision avoidance,'' in \emph{ISRR}, 2011.

\bibitem{deitke2022retrospectives}
M.~Deitke, D.~Batra, Y.~Bisk, T.~Campari, A.~X. Chang, D.~S. Chaplot, C.~Chen, C.~P. D'Arpino, K.~Ehsani, A.~Farhadi, \emph{et~al.}, ``Retrospectives on the embodied ai workshop,'' \emph{arXiv preprint arXiv:2210.06849}, 2022.

\bibitem{coumans2019}
E.~Coumans and Y.~Bai, ``Pybullet, a python module for physics simulation for games, robotics and machine learning,'' 2016, accessed: April 25th 2023. [Online]. Available: \url{https://pybullet.org/wordpress/}.

\bibitem{briceno2018makehuman}
L.~Briceno and G.~Paul, ``Makehuman: A review of the modelling framework,'' in \emph{IEA}, 2018.

\bibitem{deloitte2010employment}
D.~J. Deloitte, ``Employment densities guide,'' 2010, accessed: May 29th 2023. [Online]. Available: \url{https://assets.publishing.service.gov.uk/government/uploads/system/uploads/attachment_data/file/378203/employ-den.pdf}.

\bibitem{van2020generalized}
W.~van Toll, F.~Grzeskowiak, A.~L. Gand{\'\i}a, J.~Amirian, F.~Berton, J.~Bruneau, B.~C. Daniel, A.~Jovane, and J.~Pettr{\'e}, ``Generalized microscropic crowd simulation using costs in velocity space,'' in \emph{Symposium on Interactive 3D Graphics and Games}, 2020.

\bibitem{tolani2021visual}
V.~Tolani and et.al., ``Visual navigation among humans with optimal control as a supervisor,'' \emph{RA-L}, 2021.

\bibitem{geng2023human}
Z.~Geng, C.~Wang, Y.~Wei, Z.~Liu, H.~Li, and H.~Hu, ``Human pose as compositional tokens,'' in \emph{CVPR}, 2023.

\bibitem{anderson2018evaluation}
P.~Anderson, A.~Chang, D.~S. Chaplot, A.~Dosovitskiy, S.~Gupta, V.~Koltun, J.~Kosecka, J.~Malik, R.~Mottaghi, M.~Savva, \emph{et~al.}, ``On evaluation of embodied navigation agents,'' \emph{arXiv preprint arXiv:1807.06757}, 2018.

\bibitem{ramakrishnan2022poni}
S.~K. Ramakrishnan, D.~S. Chaplot, Z.~Al-Halah, J.~Malik, and K.~Grauman, ``Poni: Potential functions for objectgoal navigation with interaction-free learning,'' in \emph{CVPR}, 2022.

\bibitem{chen2022ros}
G.~Chen, H.~Yang, and I.~M. Mitchell, ``Ros-x-habitat: Bridging the ros ecosystem with embodied ai,'' in \emph{CRV}, 2022.

\bibitem{rosano2021embodied}
M.~Rosano, A.~Furnari, L.~Gulino, and G.~M. Farinella, ``On embodied visual navigation in real environments through habitat,'' in \emph{ICPR}, 2021.

\end{thebibliography}
   
\end{document}